\documentclass[nohyperref]{article}

\usepackage{microtype}
\usepackage{graphicx}
\usepackage{subfigure}
\usepackage{booktabs}
\usepackage{multirow}
\usepackage{hyperref}
\usepackage{tipa}

\usepackage[accepted]{icml2022}

\usepackage{amsmath}
\usepackage{amssymb}
\usepackage{mathtools}
\usepackage{amsthm}
\usepackage{marvosym}
\usepackage{setspace}

\usepackage[capitalize,noabbrev]{cleveref}

\theoremstyle{plain}

\theoremstyle{definition}

\theoremstyle{remark}

\usepackage[textsize=tiny]{todonotes}

\renewcommand{\algorithmicrequire}{\textbf{Input:}}
\renewcommand{\algorithmicensure}{\textbf{Layer params:}}

\newcommand{\update}[1]{{\textcolor{black}{#1}}}

\icmltitlerunning{Flowformer: Linearizing Transformers with Conservation Flows}

\begin{document}

\twocolumn[
\icmltitle{Flowformer: Linearizing Transformers with Conservation Flows}

\icmlsetsymbol{equal}{*}

\begin{icmlauthorlist}
\icmlauthor{Haixu Wu}{software}
\icmlauthor{Jialong Wu}{software}
\icmlauthor{Jiehui Xu}{software}
\icmlauthor{Jianmin Wang}{software}
\icmlauthor{Mingsheng Long}{software}
\end{icmlauthorlist}

\icmlaffiliation{software}{School of Software, BNRist, Tsinghua University.
Haixu Wu $<$whx20@mails.tsinghua.edu.cn$>$}
\icmlcorrespondingauthor{Mingsheng Long}{mingsheng@tsinghua.edu.cn}

\icmlkeywords{Deep Learning, Efficient Transformers, Conservation Flow, Flow-Attention}

\vskip 0.3in
]

\printAffiliationsAndNotice{}  

\begin{abstract}
Transformers based on the attention mechanism have achieved impressive success in various areas. 
However, the attention mechanism has a quadratic complexity, significantly impeding Transformers from dealing with numerous tokens and scaling up to bigger models. 
Previous methods mainly utilize the similarity decomposition and the associativity of matrix multiplication to devise linear-time attention mechanisms. 
They avoid degeneration of attention to a trivial distribution by reintroducing inductive biases such as the locality, thereby at the expense of model generality and expressiveness. 
In this paper, we linearize Transformers free from specific inductive biases based on the flow network theory.
We cast attention as the information flow aggregated from the sources (values) to the sinks (results) through the learned flow capacities (attentions). 
Within this framework, we apply the property of \emph{flow conservation} into attention and propose the Flow-Attention mechanism of linear complexity.
By respectively conserving the incoming flow of sinks for \emph{source competition} and the outgoing flow of sources for \emph{sink allocation}, Flow-Attention inherently generates informative attentions without using specific inductive biases. 
Empowered by the Flow-Attention, Flowformer yields strong performance in linear time for wide areas, including long sequence, time series, vision, natural language, and reinforcement learning. 
The code and settings are available at this repository: \href{https://github.com/thuml/Flowformer}{https://github.com/thuml/Flowformer}.
\end{abstract}

\section{Introduction}
\label{Introduction}

Recently, Transformers \citep{NIPS2017_3f5ee243} have shown immense capability in sequential modeling and been widely used in various areas, such as natural language processing \citep{Devlin2019BERTPO,liu2019roberta,NEURIPS2020_1457c0d6}, computer vision \citep{dosovitskiy2021an,liu2021Swin}, time series analysis \citep{haoyietal-informer-2021,wu2021autoformer} and reinforcement learning \citep{chen2021decisiontransformer,janner2021sequence}. Based on attention mechanisms, Transformers can learn the relation between each pair of tokens in a sequence. 

However, suffering from the quadratic complexity of pairwise relation modeling, it is computationally prohibitive for Transformers to deal with long sequences and scale up to bigger models. To tackle this essential obstacle for foundation models \citep{Bommasani2021OnTO}, efficient and linear Transformers have been explored. One category of methods attempts to utilize the sparsity to reduce the model captured relations \cite{Child2019GeneratingLS,Vyas2020FastTW,Zaheer2020BigBT}. By substituting the dense matrix to a sparse version, these models can obtain a lower complexity but inevitably sacrifice some valuable information, leading to the trade-off dilemma between efficiency and performance. Another mainstream category tries to abandon the computation-consuming query-key multiplication in the attention mechanism. The typical method is to substitute or approximate the softmax-based similarity in Transformers. For example, Linear Transformer \cite{Katharopoulos2020TransformersAR} introduces the decomposition method for similarity calculation and further bypasses the query-key multiplication through the associativity of matrix multiplication. However, without using the softmax function, these methods cannot guarantee the distinguishability of attention. This may result in near-uniform attention of each token to all other tokens, namely the degenerated attention, which damages the effectiveness of the attention mechanism. Although some works try to incorporate the concentration property to avoid the trivial attention \cite{Luo2021StableFA,anonymous2022cosformer}, they have to reintroduce specific inductive biases to Transformers, such as the locality in sequence, sacrificing the model generality. Thus, \emph{how to simultaneously obtain the non-trivial attention and maintain the generality as the canonical attention} is the key challenge in the advance of linearizing Transformers.

\update{Previous works demonstrate that the softmax function is essential to avoid the trivial attention \cite{performer,Peng2021RandomFA}. It is well-known that the softmax function is originally proposed as a differentiable generalization of the ``\emph{winner-take-all}'' picking maximum operation \cite{Bridle1989TrainingSM}. Thus, the softmax function can introduce the \emph{competition} among tokens in the attention mechanism, enforcing higher attention only to the essential tokens and thereby avoiding near-uniform attention weights. Based on this insight, it is a natural solution to empower transformers with built-in competition property to generate informative attention that guarantees the modeling capability.} However, the competition mechanism is irrealizable for linear Transformers because the attention weights to compete will incur the quadratic complexity. To tackle the aforementioned problems, we attempt to reconstruct the attention mechanism from a new view of flow network \cite{Ahuja1993NetworkF}, where the competition property is naturally achieved. Note that a flow network is a directed graph with information flows from one node to another under the constraint of flow capacity. Correspondingly, the attention mechanism can be reformulated as aggregating the information from sources (i.e.,~values) to sinks (i.e.,~results) through the learned flow capacities (i.e.,~attentions). We further find that by conserving the incoming flow capacity for each sink, the outgoing flow capacities of sources will compete with each other. And by conserving the outgoing flow capacity of sources, we can also obtain the competed incoming flow capacities of sinks. Thus, benefiting from the \emph{flow conservation} in flow network, the competition mechanism can be accomplished without specific inductive biases.

Based on the above insights, we introduce the \emph{flow conservation} to the attention mechanism and further propose the \emph{Flow-Attention} mechanism, which can avoid the trivial attention and simultaneously be free from specific inductive biases. Technically, by conserving the incoming flow of sinks (i.e.,~results), the \emph{source competition} mechanism is accomplished and then applied for the non-trivial information aggregation. After the information aggregation, the \emph{sink allocation} mechanism is obtained by conserving the outgoing flow of sources (i.e.,~values) and then applied to filter the aggregated information. Empowered by the Flow-Attention, Flowformer in linear complexity achieves competitive or better performance as the canonical Transformer in extensive areas.
The contributions are summarized as follows:
\begin{itemize}
    \vspace{-5pt}
  \item This paper analyzes the attention mechanism from the new view of the flow network. By introducing the \emph{flow conservation} to both the source and sink aspects, the competition among tokens is naturally achieved.
  \vspace{-3pt}
  \item Based on flow conservation, we propose the \emph{Flow-Attention} with \emph{source competition} and \emph{sink allocation} mechanisms, which can avoid degenerated attentions without incorporating specific inductive biases.
  \vspace{-3pt}
  \item Empowered by Flow-Attention, our proposed \emph{Flowformer} yields strong performance in linear time on five benchmarks, covering wide areas: long sequence, language, vision, time series and reinforcement learning.
\end{itemize}

\section{Preliminaries}
\vspace{-3pt}
\subsection{General View of Attention Mechanism}
\vspace{-3pt}

The attention mechanism is the key component of Transformers \citep{NIPS2017_3f5ee243}, which can be used to explore the underlying relations among tokens and adaptively aggregate valuable information. The input of attention mechanism contains three parts: queries $\mathbf{Q}\in\mathbb{R}^{n\times d}$, keys $\mathbf{K}\in\mathbb{R}^{m\times d}$ and values $\mathbf{V}\in\mathbb{R}^{m\times d}$. The $i$-th row of the result  $\mathbf{R}$ of the attention mechanism can be calculated as follows:
\begin{equation}\label{equ:vanilla}
	\begin{split}
		\mathbf{R}_{i}&=\sum_{j=1}^{m}\frac{S(\mathbf{Q}_{i},\mathbf{K}_{j})}{\sum_{j'=1}^{m}S(\mathbf{Q}_{i},\mathbf{K}_{j'})}\mathbf{V}_{j},\ i\in\{1,\cdots,n\},\\
	\end{split}
\end{equation}
where ${\ast}_{i}$ denotes the $i$-th row of matrix $\ast$. $S(\mathbf{Q}_{i},\mathbf{K}_{j})$ calculates the similarity between queries and keys. Thus, the attention mechanism is to aggregate the information from values based on the attention map calculated from queries and keys. In the canonical Transformer \cite{NIPS2017_3f5ee243}, $S(\mathbf{Q}_{i},\mathbf{K}_{j})$ is set as $\text{exp}(\mathbf{Q}_{i}\mathbf{K}_{j}^{\sf T})$, corresponding to the softmax function. The softmax function introduces the competition between similarity weights, which is essential to obtain a non-trivial attention. However, because of the calculation of $\mathbf{Q}\mathbf{K}^{\sf T}$, the computation complexity of Eq.~\eqref{equ:vanilla} in vanilla Transformer is quadratic in the sequence length, concretely $\mathcal{O}(nmd)$, resulting in the core limitation.

\vspace{-5pt}
\subsection{Efficient and Linear Transformers}

To break through the complexity limitation of the canonical attention mechanism, efficient and linear Transformers are widely explored. Categorized by the operation to the attention map, the paradigms roughly involve the similarity-decomposition and attention-sparsification methods.

\vspace{-5pt}
\paragraph{Similarity-decomposition methods} 
It is notable that the quadratic complexity of canonical attention is caused by the calculation of $\mathbf{Q}\mathbf{K}^{\sf T}$. This computation-consuming operation is indispensable because of the exponential definition of $S(\cdot,\cdot)$, while similarity-decomposition methods try to linearize the attention by utilizing the decomposition of $S(\cdot,\cdot)$ and the associativity of matrix multiplication. Technically, if $S(\cdot,\cdot)$ can be decomposed as the inner-product between the non-linear projections $\phi(\cdot)$ of queries and keys,
\begin{equation}
	\begin{split}\label{equ:kernel}
		S(\mathbf{Q}_{i},\mathbf{K}_{j})=\left<\phi(\mathbf{Q}_{i}), \phi(\mathbf{K}_{j})\right>=\phi(\mathbf{Q}_{i})\phi(\mathbf{K}_{j})^{\sf T},\\
	\end{split}
\end{equation}
then we can adopt associativity to reduce complexity. Under this condition, Eq.~\eqref{equ:vanilla} will be reformulated as follows:
\begin{equation}
	\begin{split}\label{equ:assoc}
		\mathbf{R}_{i}&=\sum_{j=1}^{m}\frac{\phi(\mathbf{Q}_{i})\phi(\mathbf{K}_{j})^{\sf T}}{\sum_{j'=1}^{m}\phi(\mathbf{Q}_{i})\phi(\mathbf{K}_{j'})^{\sf T}}\mathbf{V}_{j}\\
		&=\frac{\phi(\mathbf{Q}_{i})\sum_{j=1}^{m}\phi(\mathbf{K}_{j})^{\sf T}\mathbf{V}_{j}}{\phi(\mathbf{Q}_{i})\sum_{j=1}^{m}\phi(\mathbf{K}_{j})^{\sf T}},\ i\in\{1,\cdots,n\},\\
	\end{split}
\end{equation}
in which the direct calculation of $\mathbf{Q}\mathbf{K}^{\sf T}$ is avoided and replaced by the multiplication of keys and values, namely $\phi(\mathbf{K})^{\sf T}\mathbf{V}$. Correspondingly, the complexity is $\mathcal{O}(nd^2)$ and linear in sequence length. One typical instance proposed by Linear Transformer \cite{Katharopoulos2020TransformersAR} is to set the non-linear projection $\phi(\cdot)$ as $\text{elu}(\cdot)+1$ using the exponential linear unit. However, it is hard for Linear Transformer to avoid degenerated attention without the softmax function. Thus, RFA \cite{Peng2021RandomFA} and Performer \cite{performer} adopt the random Fourier features \cite{Rahimi2007RandomFF} and positive random features to approximate the softmax respectively. But these two methods suffer from the approximate error and have to choose specific kernels to meet the theoretical guarantee of approximation. \update{Besides, SOFT \cite{Lu2021SOFTST}, YOSO \cite{zeng2021yoso} and Nystr{\"o}mformer \cite{Xiong2021NystrmformerAN} approximate softmax function by Nystr{\"o}m method, while they will bring tedious iterations into calculation and cannot implement the causal version of attention for autoregressive tasks.} Recently, cosFormer \cite{anonymous2022cosformer} decomposes the similarity metric based on the decomposition of cosine function, where $S(\mathbf{Q}_{i},\mathbf{K}_{j})$ is set as $\phi(\mathbf{Q}_{i})\phi(\mathbf{K}_{j})^{\sf T}\cos(\frac{\pi}{2}\times\frac{i-j}{M})$ and $M$ is a hyperparameter. Although it can concentrate the learned attention, the design in cosFormer explicitly includes the locality inductive bias in temporal dimension, overlooking the spatial position in vision, thereby sacrificing the generality.

\vspace{-10pt}
\paragraph{Attention-sparsification methods} 
This paradigm does not change the similarity metric $S(\cdot, \cdot)$ in the canonical attention but attempts to reduce the model captured relations. Typically, Sparse Transformer \cite{Child2019GeneratingLS} only calculates the similarity between pre-selected query-key pairs, thereby able to obtain the sparse attention matrix for efficiency. Based on the low-rank hypothesis, Linformer \cite{Wang2020LinformerSW} adopts the projector to map the queries and keys to a low dimension. Reformer \cite{kitaev2020reformer} replaces the dense dot-product attention with the locality-sensitive hashing for similarity calculation. Clustered attention \cite{Vyas2020FastTW} reduces the complexity by grouping the queries. BigBird \cite{Zaheer2020BigBT} enhances the sparse attention mechanism with the global token to accomplish the more powerful information aggregation. Note that all the above methods sacrifice information utilization. Thus, they have to suffer the efficiency-performance dilemma. 

Unlike prior methods, Flowformer adopts the similarity decomposition and brings the flow conservation into design, which can naturally achieve the competition among tokens to avoid trivial attentions. Thus, Flowformer escapes from the efficiency-performance dilemma and eliminates specific inductive biases, thereby empowering stronger generality.

\section{Method}

As aforementioned, the core of linearizing Transformers is to obtain the non-trivial attention and maintain the generality simultaneously. To break with the above challenges, we reanalyze the attention mechanism from a novel view of the flow network. Inspired by the flow network theory, we propose the \emph{Flow-Attention} mechanism by conducting the flow conservation on both the source and sink aspects. This design can bring the competition mechanism to sources and the allocation mechanism to sinks, avoiding trivial attentions without incorporating specific inductive biases.

\subsection{Attention Mechanism: A Flow Network View}

From a new perspective, this paper attempts to reformulate the attention mechanism from the view of flow network \cite{Ahuja1993NetworkF}. As stated in Eq.~\eqref{equ:vanilla}, the canonical attention mechanism presents the procedure that the information is aggregated from the values $\mathbf{V}$ to the results $\mathbf{R}$. And the aggregation weights (i.e.,~attention map) are calculated from the similarity between queries $\mathbf{Q}$ and keys $\mathbf{K}$. 

\begin{figure}[h]
    \begin{center}
    \centerline{\includegraphics[width=\columnwidth]{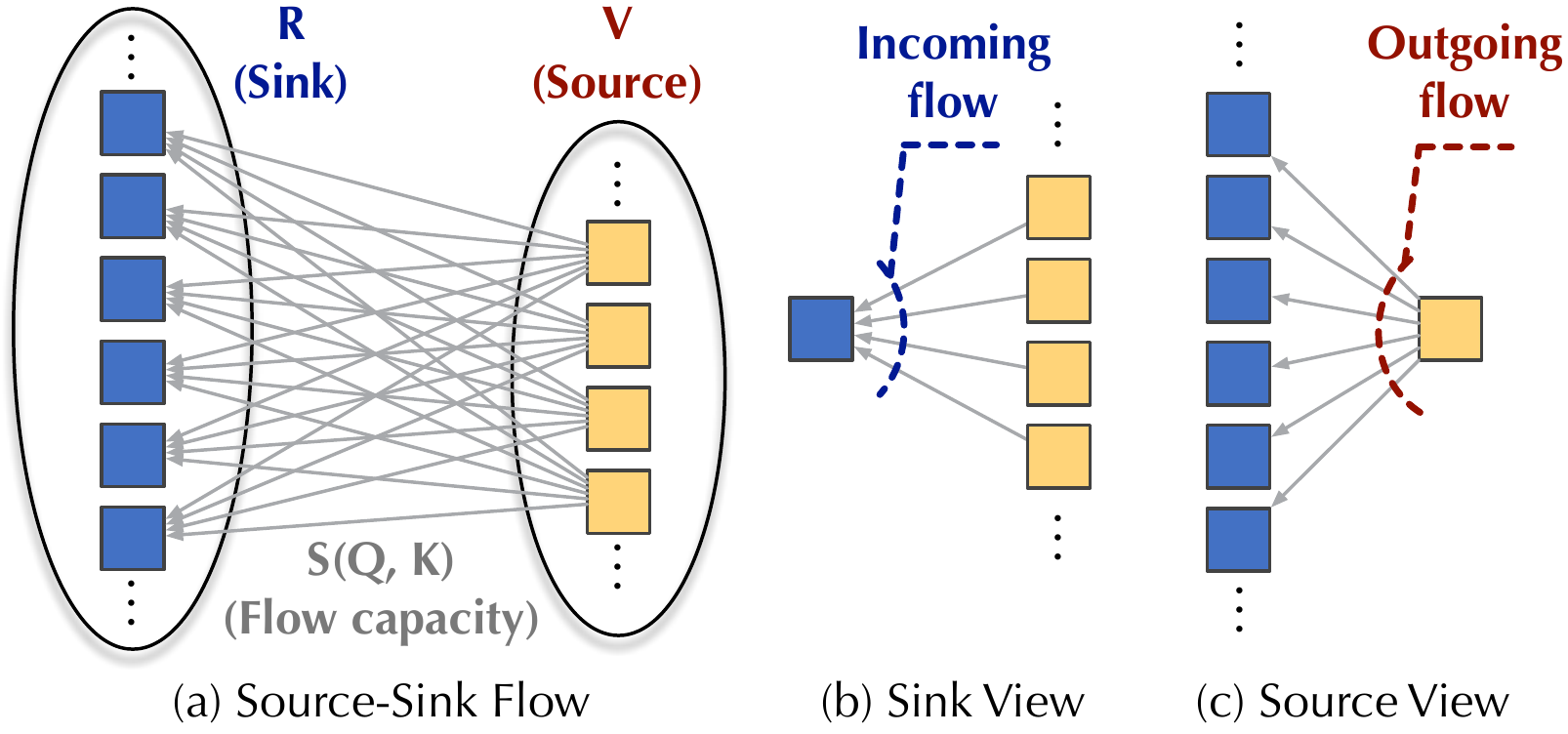}}
    \vspace{-5pt}
    \caption{The flow network view for attention. The \textcolor{blue}{blue} boxes (sinks) represent the results $\mathbf{R}$. The \textcolor{orange}{orange} boxes (sources) represent the values $\mathbf{V}$. The \textcolor{gray}{gray} arrows (flow capacity) denote the attention weights calculated from queries $\mathbf{Q}$ and keys $\mathbf{K}$.}
    \label{fig:flow-view}
    \end{center}
    \vspace{-15pt}
\end{figure}

\update{Correspondingly, we can interpret the attention mechanism from the flow network view, as shown in Figure \ref{fig:flow-view}. Here, we take the results $\mathbf{R}$ as sinks, which have only incoming information flow for source aggregation, and the values $\mathbf{V}$ as sources, which have only outgoing flow for providing information to sinks.} Following the similarity-decomposition method (Eq.~\eqref{equ:kernel}--\eqref{equ:assoc}), we can define the similarity function $S(\mathbf{Q}, \mathbf{K})$ as $\phi(\mathbf{Q})\phi(\mathbf{K})^{\sf T}$ to achieve linear complexity, where $\phi(\cdot)$ is the element-wise non-linear projection. Due to the property of flow network, we choose $\phi(\cdot)$ as a non-negative function to keep flow capacity positive.

\begin{figure*}[t]
\begin{center}
	\centerline{\includegraphics[width=1.0\textwidth]{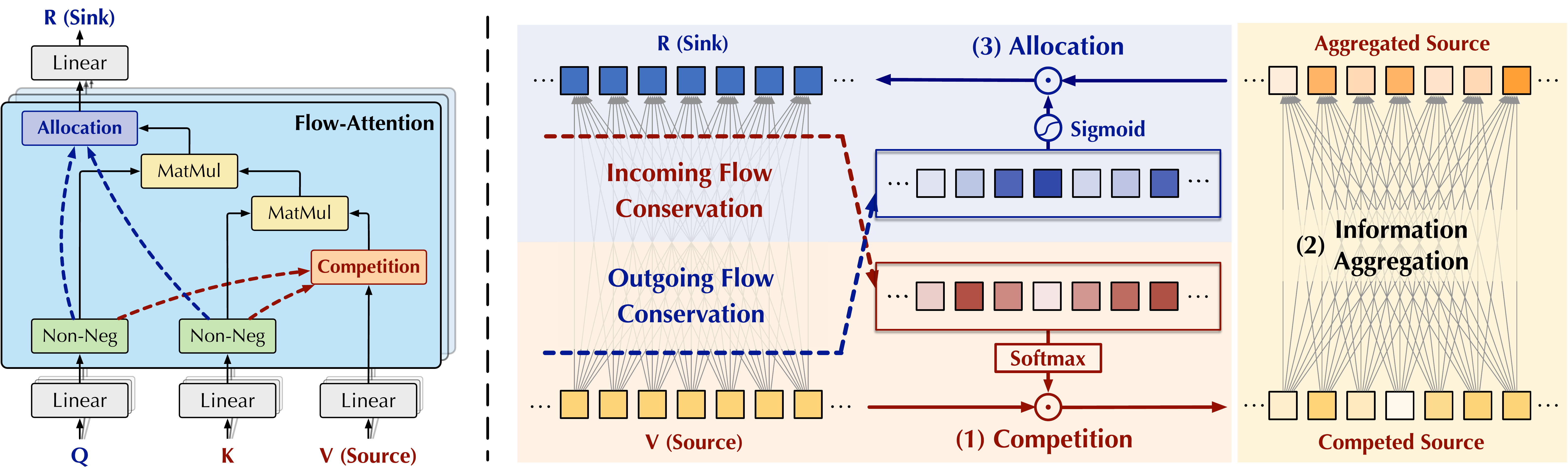}}
	\caption{The overall procedure of Flow-Attention. The \emph{source competition} mechanism (\textcolor{red}{red} dotted line) is obtained by incoming flow conservation to sinks. The \emph{sink allocation} mechanism (\textcolor{blue}{blue} dotted line) is accomplished by outgoing flow conservation to sources. }
	\label{fig:overall}
\end{center}
\vspace{-5pt}
\end{figure*}

As an important concept in flow network, the incoming and outgoing flow of each node can reflect the global interaction between each node and the whole flow network, which can provide valuable global information. Under the attention mechanism context, the flow capacity is calculated by queries and keys. Suppose we have $n$ sinks and $m$ sources. The incoming flow ${I}_{i}\in\mathbb{R}$ of the $i$-th sink and the outgoing flow ${O}_{j}\in\mathbb{R}$ of the $j$-th source can be calculated as:
\begin{equation}
	\begin{split}\label{equ:in-out-flow}
	    {I}_{i} = \phi(\mathbf{Q}_{i})\sum_{j=1}^{m}\phi(\mathbf{K}_{j})^{\sf T},\ {O}_{j} =\phi(\mathbf{K}_{j})\sum_{i=1}^{n}\phi(\mathbf{Q}_{i})^{\sf T},
	\end{split}
\end{equation}
where $i\in\{1,\cdots,n\}$ and $j\in\{1,\cdots,m\}$. Note that the incoming flow $\mathbf{I}\in\mathbb{R}^{n\times1}$ and the outgoing flow $\mathbf{O}\in\mathbb{R}^{m\times1}$ can be calculated in linear complexity by Eq.~\eqref{equ:in-out-flow}. Then, we derive the Flow-Attention mechanism based on $\mathbf{I}$ and $\mathbf{O}$.

\subsection{Flow-Attention Mechanism}

\begin{figure}[ht]
    \begin{center}
    \centerline{\includegraphics[width=\columnwidth]{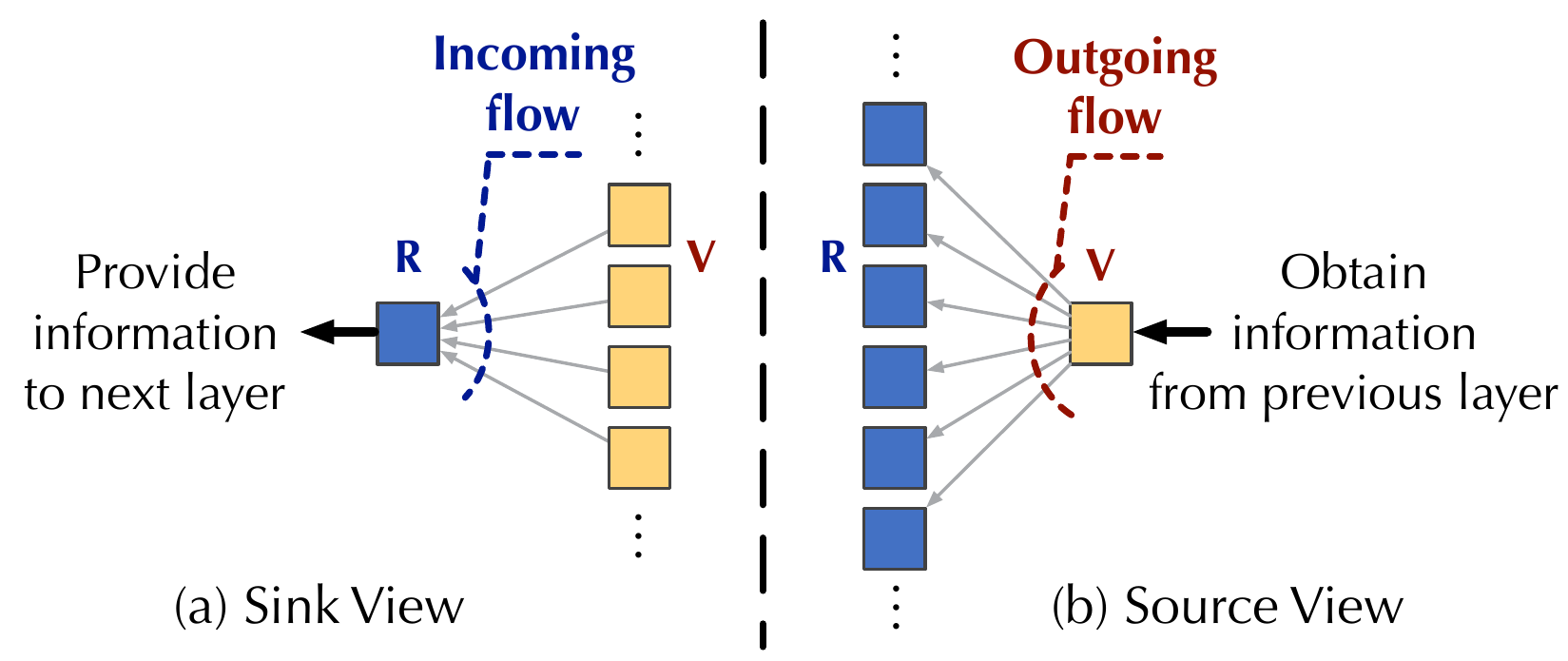}}
    \vspace{-5pt}
    \caption{\update{Information flow outside the attention mechanism.}}
    \label{fig:flow_conservation}
    \vspace{-10pt}
    \end{center}
\end{figure}

Recall that the softmax function can obtain non-trivial attention by introducing \emph{competition} among tokens, which is the indispensable component for avoiding the trivial attention. However, the competition mechanism is irrealizable for linear Transformers because the attention weights to compete will incur quadratic complexity. To tackle this dilemma, we propose the Flow-Attention by introducing the flow conservation into design in the spirit that ``fixed resource will cause competition''. \update{Intuitively, as shown in Figure \ref{fig:flow_conservation}, outside the attention mechanism, the information of values ($\mathbf{V}$) is obtained from previous layer and the information of results ($\mathbf{R}$) will be provided to the next layer. Without loss of generality, we set the information incoming from or outgoing to other layers as the default value 1 for \emph{fixing} the resource.}

\update{In the deep model implementation}, as shown in Figure \ref{fig:overall}, Flow-Attention adopts the non-negative and non-linear projection $\phi$ for the computation of flow capacity. Inspired by the flow network theory, we find that the competition mechanism is naturally achieved by introducing the \emph{conservation} property in both source and sink aspects. \update{Specifically, by conserving the incoming flow capacity for each sink as the default value 1, \emph{i.e.}~fixing the information provided to next layer (Figure \ref{fig:flow_conservation}), the outgoing flow capacities of sources will compete with each other since their sum is constrained as 1. Similarly, by conserving the outgoing flow capacity for each source as the default value 1, \emph{i.e.}~fixing the information obtained from previous layer (Figure \ref{fig:flow_conservation}), we can also obtain the competed incoming flow capacities of sinks. The above two conservation processes can be achieved by the following normalizing operations, which can well calibrate the subsequently calculated flow capacities:}
\begin{equation}
	\begin{split}\label{equ:reweighting}
	    \frac{\phi(\mathbf{K})}{\mathbf{O}}, \frac{\phi(\mathbf{Q})}{\mathbf{I}},
	\end{split}
\end{equation}
\update{where the ratio represents the element-wise division, $\frac{\phi(\mathbf{K})}{\mathbf{O}}$ is for the source conservation and $\frac{\phi(\mathbf{Q})}{\mathbf{I}}$ is for the sink conservation.
Through the normalization, the conservation of flow capacity for each source and sink token are guaranteed, which can be verified by the following equations:}
\begin{equation}
	\begin{split}\label{equ:conservation_proof}
	    \text{source-$j$:}\ & \frac{\phi(\mathbf{K}_{j})^{\sf T}}{{O}_{j}}\sum_{i=1}^{n}\phi(\mathbf{Q}_{i})=\frac{\sum_{i=1}^{n}\phi(\mathbf{Q}_{i})\phi(\mathbf{K}_{j})^{\sf T}}{{O}_{j}}=1 \\
	    \text{sink-$i$:}\ &
	    \frac{\phi(\mathbf{Q}_{i})^{\sf T}}{{I}_{i}}\sum_{j=1}^{m}\phi(\mathbf{K}_{j})=\frac{\sum_{j=1}^{m}\phi(\mathbf{K}_{j})\phi(\mathbf{Q}_{i})^{\sf T}}{{I}_{i}}=1\\
	\end{split}
\end{equation}
\update{which follows the same calculation as Eq.~\eqref{equ:in-out-flow}. The first equation is for the outgoing flow capacity of $j$-th source after the normalization $\frac{\phi(\mathbf{K})}{\mathbf{O}}$ and the second equation is for the incoming flow capacity of $i$-th sink after the normalization $\frac{\phi(\mathbf{Q})}{\mathbf{I}}$. Both capacities are equal to the default value 1.}

\update{After the conservation processes in both source and sink aspects, the competition is realized among the incoming flow of sink tokens and the outgoing flow among source tokens respectively. The conserved information flows are:}
\begin{equation}
	\begin{split}\label{equ:conservation}
	    \widehat{\mathbf{I}}= \phi(\mathbf{Q})\sum_{j=1}^{m}\frac{\phi(\mathbf{K}_{j})^{\sf T}}{{O}_{j}},\ \widehat{\mathbf{O}}= \phi(\mathbf{K})\sum_{i=1}^{n}\frac{\phi(\mathbf{Q}_{i})^{\sf T}}{{I}_{i}},
	\end{split}
\end{equation}
where $\widehat{\mathbf{I}}\in\mathbb{R}^{n\times1}$ and $\widehat{\mathbf{O}}\in\mathbb{R}^{m\times1}$ denote the capacity of conserved incoming flow and outgoing flow respectively. 

Benefiting from the competition brought by the incoming flow conservation of sinks, $\widehat{\mathbf{O}}$ denotes the information provided by the sources under the fixed sum of flow capacity, which indicates the importance of each source. As for $\widehat{\mathbf{I}}$, it denotes the information obtained by each sink when the source outgoing capacity is fixed as 1, which reflects the capacity of aggregated information that each sink is allocated. Thus, as shown in Figure \ref{fig:overall}, we present the Flow-Attention based on the above conserved information flows, which includes the competition mechanism for sources and the allocation mechanism for sinks. The overall equations of the Flow-Attention mechanism can be formalized as follows:
\begin{equation}
	\begin{split}\label{equ:overall}
	    \text{Competition:}\ &\widehat{\mathbf{V}}=\operatorname{Softmax}(\widehat{\mathbf{O}})\odot\mathbf{V}\\
		\text{Aggregation:}\ &\mathbf{A}=\frac{\phi(\mathbf{Q})}{\mathbf{I}}\big(\phi(\mathbf{K})^{\sf T}\widehat{\mathbf{V}}\big)\\
		\text{Allocation:}\ &\mathbf{R}=\operatorname{Sigmoid}(\widehat{\mathbf{I}})\odot \mathbf{A},\\
	\end{split}
\end{equation}
where $\odot$ denotes the element-wise multiplication. Note that both Softmax and Sigmoid can be computed in linear time. $\widehat{\mathbf{V}}\in\mathbb{R}^{m\times d}$ represents the competed sources, which is non-trivially re-weighted based on the incoming flow conservation. ${\mathbf{A}}\in\mathbb{R}^{m\times d}$ is the aggregated source information and calculated by the associativity of matrix multiplication. After the allocation mechanism with $\widehat{\mathbf{I}}$ to filter the incoming flow capacity for each sink, we can obtain the results $\mathbf{R}\in\mathbb{R}^{n\times d}$ of Flow-Attention. With above designs, Flow-Attention involves the competition in both source and sink aspects, thereby able to avoid trivial attention efficiently. More implementation details of both the normal and causal versions are provided in Appendix \ref{appendix:details}.

Further, by replacing the attention mechanism in the Transformer family \cite{NIPS2017_3f5ee243} with Flow-Attention, we can obtain the Flowformer without changing other designs but empower previous models with linear complexity.

Note that both the competition and allocation mechanisms are conducted based on the flow capacity directly calculated from queries and keys. Thus, different from cosFormer \cite{anonymous2022cosformer} or other linear variants, Flowformer is without specific inductive bias, which empowers our model with great generality. The calculation of Eq.~\eqref{equ:in-out-flow} and \eqref{equ:conservation} is in linear complexity with respect to the sequence length. By further utilizing the associativity of matrix multiplication, Flow-Attention can be accomplished in linear complexity.

\vspace{-5pt}
\section{Experiments}

To testify the effectiveness and generality of Flowformer, we extensively experiment on five well-established benchmarks, covering long sequence modeling, language processing, computer vision, time series and reinforcement learning. As shown in Table \ref{tab:dataset_summary}, the tasks on language modeling and reinforcement learning can verify the performance of causal-version Flow-Attention. See Appendix \ref{appendix:exp_details} for more details.

\begin{table}[h]
    \vspace{-10pt}
	\caption{Summary of experiment benchmarks. }
	\label{tab:dataset_summary}
	\vskip 0.1in
	\centering
	\begin{small}
		\begin{sc}
			\renewcommand{\multirowsetup}{\centering}
			\setlength{\tabcolsep}{3.1pt}
			\scalebox{1}{
			\begin{tabular}{l|c|c|c}
				\toprule
			    Benchmarks & Task & Version & Length \\
			    \midrule
			    LRA \citeyearpar{Tay2021LongRA}  & Sequence & normal & 1000$\sim$4000 \\
			    WikiText \citeyearpar{Merity2017PointerSM} & Language & causal & 512 \\
			    ImageNet \citeyearpar{Deng2009ImageNetAL} & Vision & normal & 49$\sim$3136 \\
			    UEA \citeyearpar{Bagnall2018TheUM} & Time Series & normal & 29$\sim$1751 \\
			    D4RL \citeyearpar{Fu2020D4RLDF} & Offline RL & causal & 60 \\
				\bottomrule
			\end{tabular}}
		\end{sc}
	\end{small}
	\vspace{-5pt}
\end{table}

\subsection{Long Sequence Modeling}

\textbf{Setup.} Long-Range Arena (LRA, \citealt{Tay2021LongRA}) is a benchmark specially designed for efficient Transformers with long input sequence. It contains five different tasks: long sequence equation calculation (ListOps, \citealt{Nangia2018ListOpsAD}), byte-level text classification (Text, \citealt{Maas2011LearningWV}), document retrieval with the ACL Anthology Network (Retrieval, \citealt{Radev2013TheAA}), image classification based on the pixel sequence on CIFAR-10 (Image, \citealt{Krizhevsky2009LearningML}) and long-range spatial dependencies discovery of images (Pathfinder, \citealt{Linsley2018LearningLS}). For fair comparison, we follow the official implementation and experiment protocol of Long-Range Arena in Jax \cite{jax2018github} and replace the full attention mechanism in vanilla Transformer with Flow-Attention. All the experiments are conducted on 2 NVIDIA 2080 Ti GPUs.

\textbf{Results.} As shown in Table \ref{tab:lra}, Flowformer achieves state-of-the-art performance in both the ListOps and Retrieval tasks and competitive performance in other tasks. \update{Overall, Flowformer achieves competitive performance over previous methods (55.23$\to$56.48).} Besides, \emph{our model consistently surpasses the vanilla Transformer in all five tasks}, even though the latter adopts the full attention mechanism with quadratic complexity. In addition, we conduct the ablation study to testify the effectiveness of each module in our design. As we show, the competition and allocation mechanisms bring 1.23 (55.25$\to$56.48) and 0.9 (55.58$\to$56.48) averaged promotion respectively.

\begin{table*}[h]
	\caption{Results on the Long-Range Arena. The best result is in bold and the second best is underlined. Classification accuracy is recorded.}
	\label{tab:lra}
	\vspace{-5pt}
	\vskip 0.15in
	\centering
	\begin{small}
		\begin{sc}
			\renewcommand{\multirowsetup}{\centering}
			\setlength{\tabcolsep}{4.3pt}
			\begin{tabular}{l|ccccc|c}
				\toprule
				Model & ListOps $\uparrow$ & Text $\uparrow$ & Retrieval $\uparrow$ & Image $\uparrow$ & Pathfinder $\uparrow$ & Avg $\uparrow$ \\
				\midrule
				Local Attention \citep{tay2021long} & 15.82 & 52.98 & 53.39 & 41.46 & 66.63 & 46.06 \\
				Linear Trans. \citep{Katharopoulos2020TransformersAR} & 16.13 & \textbf{65.90} & 53.09 & 42.34 & 75.30 & 50.55 \\
				Reformer \citep{kitaev2020reformer} & 37.27 & 56.10 & 53.40 & 38.07 & 68.50 & 50.67 \\
				Sparse Trans. \citep{Child2019GeneratingLS} & 17.07 & 63.58 & 59.59 & \textbf{44.24} & 71.71 & 51.24 \\
				Sinkhorn Trans. \citep{Tay2020SparseSA} & 33.67 & 61.20 & 53.83 & 41.23 & 67.45 & 51.29 \\
				Linformer \citep{Wang2020LinformerSW} & 35.70 & 53.94 & 52.27 & 38.56 & \underline{76.34} & 51.36 \\
				Performer \citep{performer} & 18.01 & \underline{65.40} & 53.82 & 42.77 & \textbf{77.05} & 51.41 \\
				Synthesizer \citep{Tay2021SynthesizerRS} & 36.99 & 61.68 & 54.67 & 41.61 & 69.45 & 52.88 \\
				Longformer \citep{Beltagy2020LongformerTL} & 35.63 & 62.85 & 56.89 & 42.22 & 69.71 & 53.46 \\
				Transformer \citep{NIPS2017_3f5ee243} & 36.37 & 64.27 & 57.46 & 42.44 & 71.40 & 54.39 \\
				BigBird \citep{Zaheer2020BigBT} & 36.05 & 64.02 & 59.29 & 40.83 & 74.87 & 55.01 \\
				cosFormer \citep{anonymous2022cosformer} & \underline{37.90} & 63.41 & 61.36 & 43.17 & 70.33 & 55.23 \\
				\midrule
				\textbf{Flowformer w/o Competition} & 36.80 & 63.48 & \underline{61.66} & 42.39 & 71.90 & 55.25 \\
				\textbf{Flowformer w/o Allocation} & 37.00 & 63.78 & 61.33 & 42.52 & 73.26 & \underline{55.58} \\
				\textbf{Flowformer} & \textbf{38.70} & 64.29 & \textbf{62.24} & \underline{43.20} & 73.95 & \textbf{56.48} \\
				\bottomrule
			\end{tabular}
		\end{sc}
	\end{small}
    \vspace{-10pt}
\end{table*}

\begin{table*}[tb]
	\caption{Efficiency analysis (steps per second) on the Long-Range Arena in both inference and training phases. Experiments are conducted on 2 NVIDIA 2080 Ti GPUs. The best performance is in bold and the second is underlined. ``-'' indicates the out-of-memory situation. }
	\label{tab:lra_speed}
	\vspace{-5pt}
	\vskip 0.15in
	\centering
	\begin{small}
		\begin{sc}
			\renewcommand{\multirowsetup}{\centering}
			\setlength{\tabcolsep}{6pt}
			\begin{tabular}{l|cccccccc}
				\toprule
				Model Speed & \multicolumn{4}{c}{Inference (Steps per second) } & \multicolumn{4}{c}{Train (Steps per second) } \\	\cmidrule(lr){2-5}\cmidrule(lr){6-9}
				Sequence length & 1K & 2K & 3K  & 4K  & 1K & 2K & 3K & 4K  \\
				\midrule
				Transformer \citep{NIPS2017_3f5ee243} & 81.83& 	25.26&	-&	- & 22.12 &	7.50&	-&	- \\
				\midrule
				Local Attention \citep{tay2021long} & 98.28 &	96.51 &	94.60	& 95.60 & 46.75 &	43.05 & 	35.42 &	30.34 \\
				Linear Trans. \citep{Katharopoulos2020TransformersAR} &97.33 & 96.14& 94.03& 	93.69& \underline{48.66}&	\textbf{48.78}&	\underline{41.66} &	35.44  \\
				Reformer \citep{kitaev2020reformer} &  60.92&	60.30&	39.37&	26.98&  46.07&	22.93&	14.34&	9.56\\
				Sparse Trans. \citep{Child2019GeneratingLS} & 78.30&	23.33 & - & - & 21.74&	7.30 &	-	 & - \\
				Sinkhorn Trans. \citep{Tay2020SparseSA} & 91.42&	92.21&	92.72&	80.67 & 45.93&	36.21&	28.11&	23.83 \\
				Linformer \citep{Wang2020LinformerSW} & 96.56 &	\textbf{96.84} &	94.74 &	93.59 & 45.57&	44.11&	37.28&	31.58 \\
				Performer \citep{performer} & \textbf{99.60} &\underline{96.80} & \textbf{96.52} & \textbf{96.42} & 47.34 &	\underline{48.30} &	41.00&	\underline{36.14} \\
				Synthesizer \citep{Tay2021SynthesizerRS} & 65.44 &	-&	-&	-&5.16&	-&	-	&- \\
				Longformer \citep{Beltagy2020LongformerTL} & 73.56&	-&	-&	- & 13.09&	-&	-&	-\\
				BigBird \citep{Zaheer2020BigBT} & 82.50&	54.12&	37.83&	29.34& 27.34&	16.95&	12.00&	9.33 \\
				cosFormer \citep{anonymous2022cosformer} & 96.46 &	95.58 &	95.19& 	94.69& 46.50&	45.24&	39.49&	35.09\\
				\midrule
				\textbf{Flowformer} & \underline{98.83} &	96.21&	\underline{95.65} &	\underline{95.82} & \textbf{49.76} & 47.18&	\textbf{41.93}&	\textbf{36.79} \\
				\bottomrule
			\end{tabular}
		\end{sc}
	\end{small}
	\vspace{-10pt}
\end{table*}

\textbf{Efficiency.} We conduct experiments on LRA to evaluate the efficiency of our model in dealing with long sequences. As shown in Table \ref{tab:lra_speed}, Flowformer presents great efficiency in both the training and inference phases under different input sequence lengths (1K$\sim$4K). Especially compared to Performer \cite{performer}, Flowformer achieves the competitive speed for inference and the better efficiency for training. Not only with comparable efficiency, Flowformer also surpasses Performer by a large margin (51.41$\to$56.48).

\subsection{Language Modeling}

\textbf{Setup.} We conduct the language modeling experiment on the WikiText-103 \cite{Merity2017PointerSM}, which is to estimate the probability distribution of a token given the previous ones. We use this task to testify the causal version of Flow-Attention. Following the well-established experiment setting \cite{Peng2021RandomFA}, the sequence length is set as 512 for both training and evaluation. The model architecture consists of 6 decoder layers with 8 heads and 512 hidden channels for attention mechanism \cite{ott2019fairseq}. The number of hidden channels for the feed-forward layers is set as 2048. \update{All the models are trained from scratch without pre-training on 4 NVIDIA TITAN RTX 24GB GPUs for 150K updates after a 6K-steps warm-up.}

\begin{table}[h]
	\caption{Results on language modeling with WikiText-103 \cite{Merity2017PointerSM}. A lower perplexity indicates the better results.}
	\label{tab:lm}
	\vskip 0.1in
	\centering
	\begin{small}
		\begin{sc}
			\renewcommand{\multirowsetup}{\centering}
			\scalebox{1}{
			\begin{tabular}{l|ccccccccc}
				\toprule
			    Model & Perplexity $\downarrow$ \\
				\midrule
                Transformer \citeyearpar{NIPS2017_3f5ee243} & 33.0 \\
                Linear Trans. \citeyearpar{Katharopoulos2020TransformersAR} & 38.4 \\
                Reformer \citeyearpar{kitaev2020reformer} & \update{33.6} \\
                Performer \citeyearpar{performer} & 37.5 \\
                TRF-Transformer \citeyearpar{Peng2021RandomFA} & 33.6 \\
                TRF-Transformer-GATE \citeyearpar{Peng2021RandomFA} & 31.3 \\
                cosFormer \citeyearpar{anonymous2022cosformer} & 34.1 \\
                \midrule
                \textbf{Flowformer w/o Competition} & 31.2 \\
                \textbf{Flowformer w/o Allocation} & 32.2 \\
                \textbf{Flowformer} & \textbf{30.8} \\
				\bottomrule
			\end{tabular}}
		\end{sc}
	\end{small}
	\vspace{-10pt}
\end{table}

\begin{figure*}[h]
\begin{center}
	\centerline{\includegraphics[width=2.0\columnwidth]{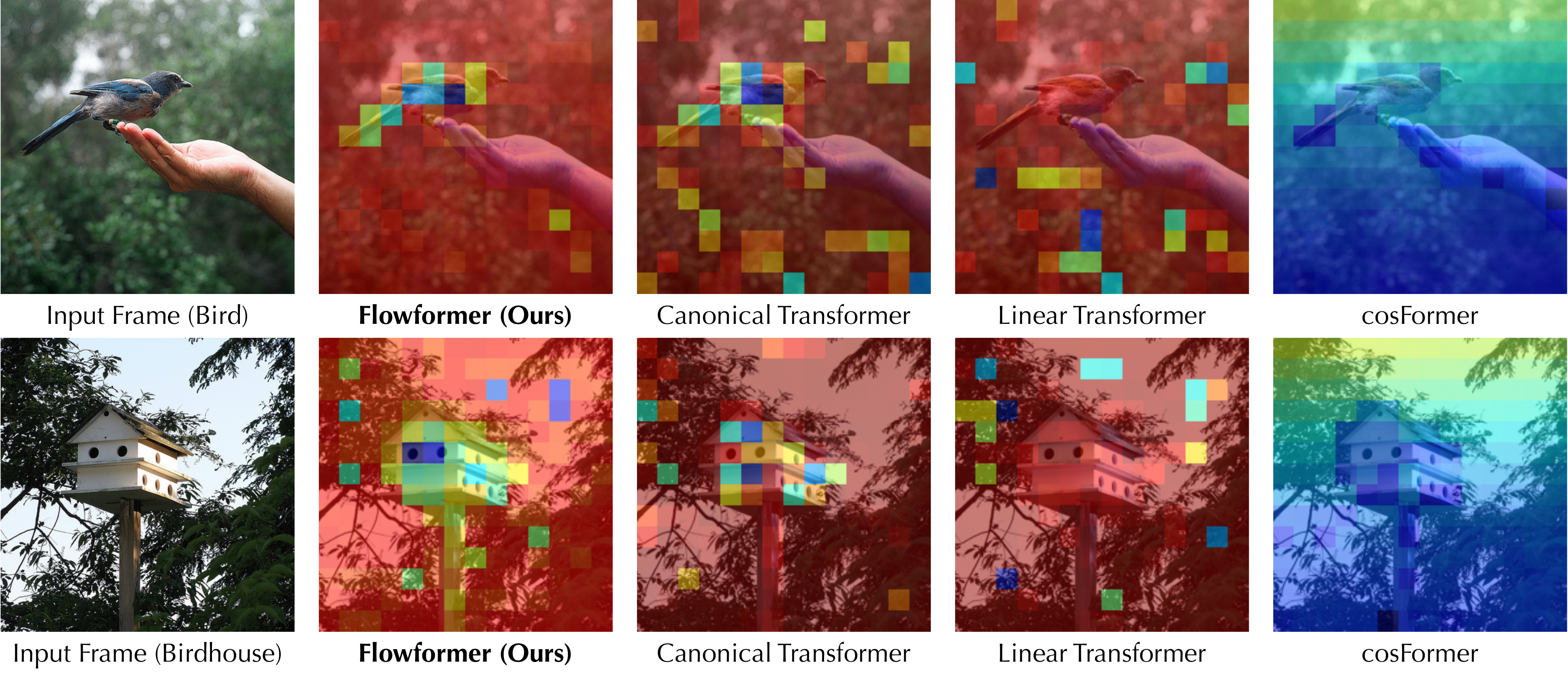}}
	\vspace{-5pt}
	\caption{Visualization of learned attention. We present the sum of attention weights to each frame patch in the last layer of the model. For Flowformer, we visualize the competition weights $\operatorname{Softmax}(\widehat{\mathbf{O}})\in\mathbb{R}^{m\times1}$, which is applied to sources for non-trivial aggregation.}
	\label{fig:attn_vis}
\end{center}
\vspace{-15pt}
\end{figure*}

\textbf{Results.} We can find that Flowformer achieves the best performance in language modeling task from Table \ref{tab:lm} and even outperforms the vanilla Transformer \cite{NIPS2017_3f5ee243}. Since the language modeling is an autoregressive task, these experiments can also verify the effectiveness of Flow-Attention in the causal version. Besides, we conduct the ablation study for both the competition and allocation mechanisms. The results prove the effectiveness of these two modules, where competition and allocation mechanisms bring 1.4 and 0.4 absolute reductions in perplexity respectively. \update{It is also notable that, both the competition and allocation mechanisms are based on flow conservation. Thus, any one of them can avoid trivial attention to some extent.}

\subsection{Image Recognition}

\textbf{Setup.} We testify the capability of Flowformer in image recognition by experimenting on the ImageNet-1K \cite{Deng2009ImageNetAL}. This dataset contains 1.28M training images and 50K validation images with 1,000 classes. Each image is in the resolution of $224\times 224$. The Top-1 accuracy and Top-5 accuracy are recorded as the metrics. \update{To fully evaluate our proposed Flowformer}, we demonstrate the experiments in the following two aspects:
\begin{itemize}
    \item Compare different attentions under the same Transformer architecture. \update{We present Flowformer with 19 layers in a four-stage hierarchical structure, where the channels are in $\{96,192,384,768\}$ and the input sequence length for each stage is in $\{3136,784,196,49\}$ correspondingly.} Global average pooling and linear projection are employed at the end of the model for classification. We take extensive efficient Transformers as baselines, as well as ViT \citeyearpar{dosovitskiy2021an}.
    \item Apply the Flow-Attention to the specific-designed vision Transformer, such as DeiT \cite{Touvron2021TrainingDI}, which adopts the token distillation for data efficiency.
\end{itemize}
All the experiments are conducted on 8 NVIDIA TITAN RTX 24GB GPUs for 300 epochs.

\begin{table}[h]
    \vspace{-5pt}
	\caption{\update{Accuracy results (\%) on ImageNet-1K \cite{Deng2009ImageNetAL}. A higher accuracy indicates the better performance.}}
	\label{tab:imagenet}
	\vskip 0.1in
	\centering
	\begin{small}
		\begin{sc}
			\renewcommand{\multirowsetup}{\centering}
			\setlength{\tabcolsep}{0.4pt}
			\scalebox{1}{
			\begin{tabular}{l|ccc|cccccc}
				\toprule
			    \scalebox{0.9}{\multirow{2}{*}{Model}} & \scalebox{0.9}{\multirow{2}{*}{Complex.}} & \scalebox{0.9}{Params} & \scalebox{0.9}{Flops} & \scalebox{0.9}{Top-1} & \scalebox{0.9}{Top-5} \\
			     &  & \scalebox{0.9}{(MB)} & \scalebox{0.9}{(G)} & \scalebox{0.9}{Acc.} & \scalebox{0.9}{Acc.} \\
				\midrule
				ViT-base \citeyearpar{dosovitskiy2021an} & \scalebox{0.9}{$\mathcal{O}(n^2d)$} & 86 & 55.4 & 77.9 & / \\
				ViT-large \citeyearpar{dosovitskiy2021an} & \scalebox{0.9}{$\mathcal{O}(n^2d)$} & 307 & 190.7 & 76.5 & / \\
				Full Attn. \citeyearpar{NIPS2017_3f5ee243} & \scalebox{0.9}{$\mathcal{O}(n^2d)$} & 41 & 6.7 & 78.7 & 94.3 \\
				Linear Trans. \citeyearpar{Katharopoulos2020TransformersAR} & \scalebox{0.9}{$\mathcal{O}(nd^2)$} & 41 & 6.3 & 79.0 & 94.1 \\
				Reformer \citeyearpar{kitaev2020reformer} &  \scalebox{0.8}{$\mathcal{O}\left((n\log n)d\right)$} & 37 & 6.0 & 79.6 & 94.7 \\
				Longformer \citeyearpar{Beltagy2020LongformerTL} & \scalebox{0.9}{$\mathcal{O}(nd^2)$} & 38 & 6.3  & 77.6 & 93.1 \\
				Performer \citeyearpar{performer} & \scalebox{0.9}{$\mathcal{O}(nd^2)$} & 41 & 6.3 & 78.1 & 93.2 \\
				\scalebox{0.85}{Nystr{\"o}mformer} \citeyearpar{Xiong2021NystrmformerAN} & \scalebox{0.9}{$\mathcal{O}(nd^2)$} & 41 & 6.3 & 77.2 & 93.0 \\
				YOSO-E \citeyearpar{zeng2021yoso} & \scalebox{0.9}{$\mathcal{O}(nd^2)$} & 41 & 5.8 & 79.0 & 94.3 \\
				SOFT \citeyearpar{Lu2021SOFTST} & \scalebox{0.9}{$\mathcal{O}(nd^2)$} & 37 & 5.8 & 79.2 & 94.5  \\
				cosFormer \citeyearpar{anonymous2022cosformer} & \scalebox{0.9}{$\mathcal{O}(nd^2)$} & 41 & 6.3 & 68.3 & 88.0 \\
			    \textbf{Flowformer} & \scalebox{0.9}{$\mathcal{O}(nd^2)$} &41 & 6.3 & \textbf{80.6} & \textbf{94.9} \\
				\midrule
				DeiT-S \citeyearpar{Touvron2021TrainingDI} & \scalebox{0.9}{$\mathcal{O}(n^2d)$} & 22 & 4.6 & 79.8 & \textbf{95.0} \\
				\textbf{DeiT+Flowformer} & \scalebox{0.9}{$\mathcal{O}(nd^2)$} & 22 & \textbf{4.2} & \textbf{80.0} & 94.8 \\
				\bottomrule
			\end{tabular}}
		\end{sc}
	\end{small}
\end{table}

\textbf{Results.} 
Table \ref{tab:imagenet} shows that Flowformer achieves the strong performance in both the Top-1 and Top-5 accuracy metrics along with the linear complexity. Besides, Flowformer is the best linear Transformer and even surpasses the vanilla Transformer (Top-1: 78.7 v.s.~80.4). It is notable that, comparing to Flowformer, cosFormer \cite{anonymous2022cosformer} provides a relatively poor result (Top-1: 68.3 v.s.~80.4). This is because that cosFormer directly incorporates the locality inductive bias along the patch sequence and does not consider the spatial position.
This experiment also demonstrates the importance of specific-inductive-bias-free design, which can be naturally satisfied in Flowformer.

\begin{table*}[tb]
	\caption{\update{Accuracy results (\%) on time series classification. A higher accuracy indicates the better performance. As for the baselines of the Transformer family, we include the the canonical Transformer (Trans.), Linear Transformer (Linear.), Reformer (Re.), Longformer (Long.), Performer (Per.), cosFormer (cos.) and etc for a comprehensive comparison. }  }
	\label{tab:tsc}
	\vspace{-5pt}
	\vskip 0.15in
	\centering
	\begin{small}
		\begin{sc}
        	\renewcommand{\multirowsetup}{\centering}
			\setlength{\tabcolsep}{0.23pt}
			\begin{tabular}{l|cccccccccccccc}
				\toprule
				\multirow{4}{*}{Dataset / Model} & \multicolumn{3}{c}{\multirow{2}{*}{Classical methods}} & \multicolumn{10}{c}{Deep Models} \\
				\cmidrule(lr){5-15}
				& & & & RNN & TCN & \multicolumn{9}{c}{Transformer and its Efficient Variants} \\
				\cmidrule(lr){2-4}\cmidrule(lr){5-5}\cmidrule(lr){6-6}\cmidrule(lr){7-15}
				 & \scalebox{0.9}{DTW} & \scalebox{0.8}{XGBoost} & \scalebox{0.9}{Rocket}  & \scalebox{0.9}{LSTM}  & \scalebox{0.8}{Unsuper.} & \scalebox{0.9}{Trans.} & \scalebox{0.9}{Linear.} & \scalebox{0.8}{Re.} & \scalebox{0.9}{Long.} & \scalebox{0.9}{Per.} &  \scalebox{0.9}{YOSO-E} & \scalebox{0.9}{SOFT} & \scalebox{0.9}{cos.} & \scalebox{0.9}{\textbf{Flow.}} \\
				& \citeyearpar{Berndt1994UsingDT} & \citeyearpar{Chen2016XGBoostAS} &  \citeyearpar{Dempster2020ROCKETEF} & \citeyearpar{Hochreiter1997LongSM} & \citeyearpar{Franceschi2019UnsupervisedSR} & \citeyearpar{NIPS2017_3f5ee243} & \citeyearpar{Katharopoulos2020TransformersAR} & 
				\citeyearpar{kitaev2020reformer} & \citeyearpar{Beltagy2020LongformerTL} & \citeyearpar{performer} &  \citeyearpar{zeng2021yoso} &  \citeyearpar{Lu2021SOFTST} & \citeyearpar{anonymous2022cosformer} & \textbf{(ours)} \\
				\midrule
				\scalebox{0.8}{EthanolConcentration} & 32.3 & 43.7 & 45.2 & 32.3 & 28.9 & 32.7 &31.9& 31.9 & 32.3 & 31.2 &  31.2 & 32.3 & 33.5 & 33.8 \\
				\scalebox{0.8}{FaceDetection} & 52.9  & 63.3 & 64.7& 57.7 & 52.8 & 67.3 & 67.0 & 68.6 & 62.6 & 67.0&  67.3 & 64.8 & 67.1 & 67.6 \\
				\scalebox{0.8}{Handwriting} & 28.6  & 15.8& 58.8 & 15.2 & 53.3 & 32.0 & 34.7 & 27.4 & 39.6 & 32.1 &  30.9 & 28.9 & 34.7 & 33.8 \\
				\scalebox{0.8}{Heartbeat} & 71.7  & 73.2& 75.6 & 72.2 & 75.6 & 76.1 & 76.6& 77.1 & 78.0 & 75.6 & 76.5 & 77.1 & 75.6 & 77.6 \\
				\scalebox{0.8}{JapaneseVowels} & 94.9  & 86.5 & 96.2& 79.7 & 98.9 & 98.7 &99.2& 97.8 &  98.9 & 98.1 & 98.6 & 98.3 & 99.2 & 98.9\\
				\scalebox{0.8}{PEMS-SF} & 71.1  & 98.3& 75.1 & 39.9 & 68.8 & 82.1 &82.1 & 82.7 & 83.8 & 80.9 &  85.2 & 83.2 & 80.9 &  83.8\\
				\scalebox{0.9}{SelfRegulationSCP1} & 77.7  & 84.6& 90.8 & 68.9 & 84.6 & 92.2 &92.5 & 90.4 & 90.1 & 91.5 & 91.1 & 91.1 & 91.8 & 92.5\\
			    \scalebox{0.8}{SelfRegulationSCP2} & 53.9 & 48.9& 53.3  & 46.6 & 55.6 &53.9 & 56.7 & 56.7 & 55.6 & 56.7 & 53.9 & 55.0 & 55.6 & 56.1\\
			    \scalebox{0.8}{SpokenArabicDigits} & 96.3  & 69.6& 71.2 & 31.9 & 95.6 & 98.4 &98.0 & 97.0 & 94.4 & 98.4 & 98.9 & 98.4 & 98.8& 98.8\\
			    \scalebox{0.8}{UWaveGestureLibrary} & 90.3 & 75.9& 94.4 & 41.2 & 88.4 & 85.6 &85.0 & 85.6 & 87.5 & 85.3 &  88.4 & 85.6 & 85.0 & 86.6\\
			    \midrule
			    \scalebox{0.95}{Average Accuracy} & 67.0 & 66.0  & \underline{72.5} &48.6 & 70.3 & 71.9 & 72.4 & 71.5 & 72.0 & 71.9 &  72.2 & 71.5 &72.2 & \textbf{73.0} \\
				\bottomrule
			\end{tabular}
		\end{sc}
	\end{small}
\end{table*}

\textbf{Efficiency.} In addition to the theoretical complexity analysis, we also provide the parameter and computation efficiency comparison in Table \ref{tab:imagenet}. Note that, compared to the canonical attention mechanism, Flow-Attention brings no extra model parameters. Thus, a promising way is to incorporate the Flow-Attention along with other well-designed vision Transformers, which can obtain comparable results with linear complexity, such as DeiT v.s.~DeiT+Flowformer (Top-1: 79.8 $\to$ 80.0, Flops: 4.6 $\to$ 4.2). Since Flowformer is linear in sequence length, with the longer patch sequence, the efficiency promotion will be more remarkable, which is important for the models to scale up.

\begin{figure}[h]
    \begin{center}
    \centerline{\includegraphics[width=\columnwidth]{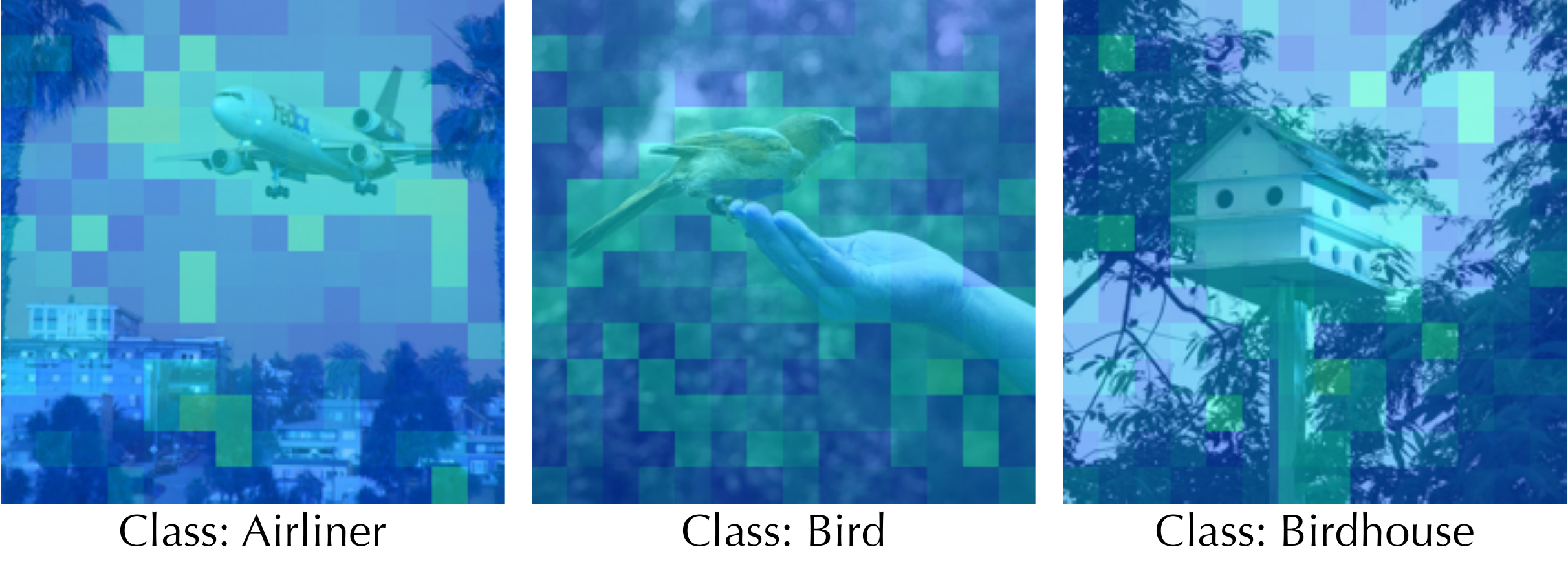}}
    \vspace{-5pt}
    \caption{Allocation weights $\operatorname{Sigmoid}(\widehat{\mathbf{I}})$ heatmap in Flowformer.}
    \label{fig:allocation_vis}
    \end{center}
    \vspace{-10pt}
\end{figure}

\textbf{Attention visualization.} To further elaborate the difference between Flowformer and other Transformers, we visualize the learned attention in Figure \ref{fig:attn_vis}. We can find that both the Flowformer and canonical Transformer \cite{NIPS2017_3f5ee243} can capture the essential parts correctly, while the latter will consume the quadratic complexity and the attention may be distracted by the background context. In contrast, without competition mechanism, Linear Transformer \cite{Katharopoulos2020TransformersAR} fails in attending to the right area and presents a degenerated attention map. As for the cosFormer \cite{anonymous2022cosformer}, due to the unsuitable introduction of sequence dimension locality and the overlook of spatial position, the attention is only concentrated on the upper part of frames, which will impede the model capability. Based on above comparisons, we can verify the advantages of Flow-Attention in informative attention learning.

\textbf{Allocation visualization.} As stated in Eq.~\eqref{equ:overall}, the learned allocation weights can reflect the capacity that each sink accepts. From Figure \ref{fig:allocation_vis}, we find that inflow capacities remain large on the essential parts, which means that these parts require more global information from sources for classification and matches our expectation.

\subsection{Time Series Classification}

\textbf{Setup.} We adopt the time series classification task to evaluate the model performance for temporal sequences. We select 10 multivariate datasets from UEA Time Series Classification Archive \cite{Bagnall2018TheUM} for experiments and follow the data pre-processing in \cite{Zerveas2021ATF}. We use 2 layers for Transformer-based models with 512 hidden channels and 8 heads for the attention mechanism. In addition to the deep models, we also compare our model with classical methods: DTW \cite{Berndt1994UsingDT}, XGBoost \cite{Chen2016XGBoostAS} and Rocket \cite{Dempster2020ROCKETEF}. Rocket is the state-of-the-art model for time series classification. All the experiments are conducted on one single NVIDIA TITAN RTX 24GB GPU for 100 epochs.

\begin{table*}[h]
	\caption{\update{Reward results on D4RL \cite{Fu2020D4RLDF}. A higher reward or a lower deviation indicates the better performance.}}
	\label{tab:rl}
	\vskip 0.1in
	\centering
	\begin{small}
		\begin{sc}
			\renewcommand{\multirowsetup}{\centering}
			\setlength{\tabcolsep}{3.5pt}
			\scalebox{1}{
			\begin{tabular}{l|ccccccc|c}
				\toprule
                \scalebox{1}{\multirow{2}{*}{Environment}} & BC & AWAC & DT & Linear Trans. & Reformer & Performer & cosFormer & \textbf{Flowformer} \\
                & \citeyearpar{pomerleau1989alvinn} & \citeyearpar{nair2020awac} & \citeyearpar{chen2021decision} &  \citeyearpar{Katharopoulos2020TransformersAR} & \citeyearpar{kitaev2020reformer} & \citeyearpar{performer} & \citeyearpar{anonymous2022cosformer} & \textbf{(Ours)} \\
				\midrule
				\multicolumn{9}{c}{Medium-Expert}\\
				\midrule
				HalfCheetah & 55.2 & 42.8 & 83.8$\pm$3.3 & 78.2$\pm$3.2 & 81.5$\pm$1.6 & 85.1$\pm$2.1 & 85.5$\pm$2.9 & 90.8$\pm$0.4 \\
				Hopper & 52.5 & 55.8 & 104.0$\pm$2.5 & 107.2$\pm$0.9 & 104.2$\pm$9.8 & 93.5$\pm$13.9 & 98.1$\pm$7.4 & 109.9$\pm$1.0 \\
				Walker & 107.5 & 74.5 & 107.7$\pm$0.6 & 67.2$\pm$27.3 & 71.4$\pm$1.8 & 72.6$\pm$2.4 & 100.5$\pm$14.5 & 108.0$\pm$0.4 \\
				\midrule
				\multicolumn{9}{c}{Medium}\\
				\midrule
				HalfCheetah & 42.6 & 43.5 & 42.4$\pm$0.1 & 42.3$\pm$0.2 & 42.2$\pm$0.1 & 42.1$\pm$0.2 & 42.1$\pm$0.3 & 42.2$\pm$0.2 \\
				Hopper & 52.9 & 57.0 & 64.2$\pm$1.1 & 58.7$\pm$0.4 &59.9$\pm$0.7 & 59.7$\pm$7.5 & 59.8$\pm$3.8 & 66.9$\pm$2.5 \\
				Walker & 75.3 & 72.4 & 70.6$\pm$3.2 & 57.9$\pm$10.6 &65.8$\pm$4.9 & 63.3$\pm$10.7 & 71.4$\pm$1.2 & 71.7$\pm$2.5 \\
				\midrule
				\multicolumn{9}{c}{Medium-Replay}\\
				\midrule
				HalfCheetah & 36.6 & 40.5 & 34.6$\pm$0.6 & 32.1$\pm$1.5 & 33.6$\pm$0.7 & 31.7$\pm$0.9 & 32.8$\pm$3.6 & 34.7$\pm$1.5 \\
				Hopper & 18.1 & 37.2 & 79.7$\pm$7.4 & 74.3$\pm$7.0 & 66.1$\pm$2.6 & 64.6$\pm$24.2 & 59.3$\pm$16.5 & 75.5$\pm$14.5 \\
				Walker & 26.0 & 27.0 & 62.9$\pm$5.0 & 62.1$\pm$7.4 & 50.1$\pm$3.5 & 61.3$\pm$6.7 & 60.5$\pm$9.9 & 62.0$\pm$3.1  \\
				\midrule
				\textbf{Avg Reward} & 51.9 & 50.1 & 72.2$\pm$\textbf{2.6} & 64.4$\pm$6.5 & 63.9$\pm$2.9 & 63.8$\pm$7.6 & 67.8$\pm$7.6 & \textbf{73.5}$\pm$2.9 \\
				\bottomrule
			\end{tabular}}
		\end{sc}
	\end{small}
\end{table*}

\textbf{Results.} Flowformer achieves the best performance on the time series classification benchmark (Table \ref{tab:tsc}). Besides other linear Transformers and deep models, Flowformer also surpasses the state-of-the-art classical methods Rocket \cite{Dempster2020ROCKETEF} and is \emph{the only method that beats classical methods on the averaged performance}. This result can verify the temporal modeling capacity of Flowformer, which is essential for sequential data. See Appendix \ref{appdix:ts_vis} for the case study in attention visualization.

\subsection{Reinforcement Learning} 

\textbf{Setup.} We consider the continuous control tasks from D4RL benchmark \cite{Fu2020D4RLDF} to evaluate the model performance on the offline reinforcement learning (Offline RL) \cite{lange2012batch, levine2020offline}. We select the HalfCheetah, Hopper and Walker as experiment environments, which are to control the movement of robot. To obtain a comprehensive evaluation, we experiment on different datasets pre-collected with three different behavior policies: Medium-Expert, Medium and Medium-Replay. Since the offline RL is an autoregressive task, it can also be used to testify the causal-version Flow-Attention. For comparison, we include the Decision Transformer (DT,~\citealt{chen2021decision}), Behavior Cloning (BC,~\citealt{pomerleau1989alvinn}), AWAC \cite{nair2020awac}, Linear Transformer (Linear Trans.,~\cite{Katharopoulos2020TransformersAR}), Reformer \cite{kitaev2020reformer}, Performer~\cite{performer} and cosFormer \cite{anonymous2022cosformer} as baselines, where DT is the state-of-the-art models for offline RL and adopts the canonical Transformer as the backbone. We adopt 3 layers with 256 hidden channels and 4 heads in all experiments for Flowformer and other Transformers. We repeat each experiment three times with different seeds on one single NVIDIA 2080 Ti GPU for 10 epochs.

\textbf{Results.} 
\update{As shown in Table \ref{tab:rl}, it is notable that compared to the vanilla Transformer used in DT, previous efficient Transformers degenerate a lot and cannot provide a stable result. Especially, the averaged rewards of Reformer \cite{kitaev2020reformer} and Performer \cite{performer} decrease seriously (72.2 v.s.~63.9 and 63.8 respectively), indicating that the locally sensitive hashing or random Fourier features may be not suitable for the global dependency modeling under the reinforcement learning context. In contrast, Flowformer still shows a competitive performance on this challenging control task in the comparison with DT (72.2 v.s.~73.5), justifying the generality of our proposed Flowformer in offline reinforcement learning.}

\section{Conclusions}

This paper focuses on Transformer linearization and attempts to tackle this problem from a new view of the flow network. By introducing the flow conservation to the attention mechanism, we present the Flow-Attention mechanism, which can naturally achieve the competition mechanism for sources and the allocation mechanism for sinks to filter the aggregated source information. Empowered by Flow-Attention, Flowformer can achieve the linear complexity and avoid degenerated attention without specific inductive biases. With great generality, Flowformer achieves the strong performance on extensive areas, covering vision, language, long sequence, time series, and reinforcement learning.

\update{Our future work includes scaling up the proposed efficient Flowformer to build general-purpose pre-trained models facilitating a wider range of upstream and downstream tasks.}

\section*{Acknowledgements}
\update{This work was supported by the National Key Research and Development Plan (2020AAA0109201), National Natural Science Foundation of China (62022050 and 62021002), Beijing Nova Program (Z201100006820041), and BNRist Innovation Fund (BNR2021RC01002).}

\bibliography{example_paper}
\bibliographystyle{icml2022}

\newpage
\appendix
\onecolumn

\section{Implementation Details}\label{appendix:details}

\subsection{Pseudo-code for Flow-Attention}

We present the pseudo-code of normal Flow-Attention in Algorithm \ref{alg:flow-attention} and the causal version in Algorithm \ref{alg:flow-attention-causal}. Especially, in the causal version, we adopt the $\texttt{Causal-Dot-Product}$ \cite{Katharopoulos2020TransformersAR} for the information aggregation.
\begin{algorithm}[hbp]
  \setstretch{1.2}
  \caption{Multi-head Flow-Attention Mechanism (normal version). }
  \label{alg:flow-attention}  
\begin{algorithmic}[1]
    \STATE {\bfseries Input:}
   $\mathbf{Q}\in\mathbb{R}^{n \times d}, \mathbf{K}\in\mathbb{R}^{m \times d}, \mathbf{V}\in\mathbb{R}^{m \times d}$ 
    \STATE{$\mathbf{Q},\mathbf{K},\mathbf{V} = \texttt{Split}(\mathbf{Q}),\texttt{Split}(\mathbf{K}),\texttt{Split}(\mathbf{V})$ } \qquad \textcolor{gray}{//\ $\mathbf{Q}\in\mathbb{R}^{n\times h\times \frac{d}{h}},\mathbf{K},\mathbf{V}\in\mathbb{R}^{m\times h\times \frac{d}{h}}$}
    \STATE{$\mathbf{Q},\mathbf{K} = \texttt{Sigmoid}(\mathbf{Q}), \texttt{Sigmoid}(\mathbf{K})$} 
    \STATE{$\mathbf{I} = \texttt{Sum}\Bigg(\mathbf{Q}\odot\texttt{Broadcast}\Big(\texttt{Sum}(\mathbf{K}, \texttt{dim=0}), \texttt{dim=0}\Big), \texttt{dim=2}\Bigg)$} \qquad \qquad \textcolor{gray}{//\ $\mathbf{I}\in\mathbb{R}^{n\times h}$}
    \STATE{$\mathbf{O} = \texttt{Sum}\Bigg(\mathbf{K}\odot\texttt{Broadcast}\Big(\texttt{Sum}(\mathbf{Q}, \texttt{dim=0}), \texttt{dim=0}\Big), \texttt{dim=2}\Bigg)$} \qquad \qquad \textcolor{gray}{//\ $\mathbf{O}\in\mathbb{R}^{m\times h}$}
    \STATE{$\widehat{\mathbf{I}} = \texttt{Sum}\Bigg(\mathbf{Q}\odot\texttt{Broadcast}\Big(\texttt{Sum}(\mathbf{K}/\mathbf{O}, \texttt{dim=0}), \texttt{dim=0}\Big), \texttt{dim=2}\Bigg)$} \qquad \qquad \textcolor{gray}{//\ $\widehat{\mathbf{I}}\in\mathbb{R}^{n\times h}$}
    \STATE{$\widehat{\mathbf{O}} = \texttt{Sum}\Bigg(\mathbf{K}\odot\texttt{Broadcast}\Big(\texttt{Sum}(\mathbf{Q}/\mathbf{I}, \texttt{dim=0}), \texttt{dim=0}\Big), \texttt{dim=2}\Bigg)$}  \qquad \qquad \textcolor{gray}{//\ $\widehat{\mathbf{O}}\in\mathbb{R}^{m\times h}$}
    \STATE {$\mathbf{R}=\texttt{Matmul}\Bigg(\mathbf{Q}/\mathbf{I},\texttt{Matmul}\Big(\mathbf{K},\mathbf{V}\odot\texttt{Softmax}(\widehat{\mathbf{O}})\Big)\Bigg)\odot\texttt{Sigmoid}(\widehat{\mathbf{I}})$} \qquad \textcolor{gray}{//\ $\mathbf{R}\in\mathbb{R}^{n\times h\times \frac{d}{h}}$}
    \STATE{\textbf{Return} $\mathbf{R}$}  
\end{algorithmic}  
\end{algorithm}

\begin{algorithm}[hbp]
  \setstretch{1.2}
  \caption{Multi-head Flow-Attention Mechanism (causal version). }
  \label{alg:flow-attention-causal}  
\begin{algorithmic}[1]
    \STATE {\bfseries Input:}
   $\mathbf{Q}\in\mathbb{R}^{n \times d}, \mathbf{K}\in\mathbb{R}^{n \times d}, \mathbf{V}\in\mathbb{R}^{n \times d}$
    \STATE{$\mathbf{D}=\texttt{Broadcast}\Big(\texttt{Arrange}(n), \texttt{dim=1}\Big)$}\qquad \textcolor{gray}{//\ $\mathbf{D}\in\mathbb{R}^{n\times h}$}
    \STATE{$\mathbf{Q},\mathbf{K},\mathbf{V} = \texttt{Split}(\mathbf{Q}),\texttt{Split}(\mathbf{K}),\texttt{Split}(\mathbf{V})$ }\qquad \textcolor{gray}{//\ $\mathbf{Q},\mathbf{K},\mathbf{V}\in\mathbb{R}^{n\times h\times \frac{d}{h}}$}
    \STATE{$\mathbf{Q},\mathbf{K} = \texttt{Sigmoid}(\mathbf{Q}), \texttt{Sigmoid}(\mathbf{K})$}
    \STATE{$\mathbf{I} = \texttt{Sum}\Big(\mathbf{Q}\odot\texttt{Cusum}(\mathbf{K}, \texttt{dim=0}),\texttt{dim=2}\Big)/\mathbf{D}$} \qquad \qquad \textcolor{gray}{//\ $\mathbf{I}\in\mathbb{R}^{n\times h}$}
    \STATE{$\mathbf{O} = \texttt{Sum}\Big(\mathbf{K}\odot\texttt{Cusum}(\mathbf{Q}, \texttt{dim=0}),\texttt{dim=2}\Big)/\mathbf{D}$} \qquad \qquad \textcolor{gray}{//\ $\mathbf{O}\in\mathbb{R}^{n\times h}$}
    \STATE{$\widehat{\mathbf{I}} = \texttt{Sum}\Big(\mathbf{Q}\odot\texttt{Cusum}(\mathbf{K}/\mathbf{O}, \texttt{dim=0}),\texttt{dim=2}\Big)/\mathbf{D}$} \qquad \qquad \textcolor{gray}{//\ $\widehat{\mathbf{I}}\in\mathbb{R}^{n\times h}$}
    \STATE{$\widehat{\mathbf{O}} = \texttt{Sum}\Big(\mathbf{K}\odot\texttt{Cusum}(\mathbf{Q}/\mathbf{I}, \texttt{dim=0}),\texttt{dim=2}\Big)/\mathbf{D}$} \qquad \qquad \textcolor{gray}{//\ $\widehat{\mathbf{O}}\in\mathbb{R}^{n\times h}$}
    \STATE{$\widehat{\mathbf{O}} = \Bigg(\texttt{Exp}(\widehat{\mathbf{O}})/\texttt{Cusum}\Big(\texttt{Exp}(\widehat{\mathbf{O}}),\texttt{dim=0}\Big)\Bigg)\odot \mathbf{D}$} \qquad \qquad \textcolor{gray}{//\ $\widehat{\mathbf{O}}\in\mathbb{R}^{n\times h}$}
    \STATE {$\mathbf{R}=\texttt{Causal-Dot-Product}\Big(\mathbf{Q}/(\mathbf{I}\odot\mathbf{D}),\mathbf{K},\mathbf{V}\odot\widehat{\mathbf{O}}\Big)\odot\texttt{Sigmoid}(\widehat{\mathbf{I}})$} \qquad \textcolor{gray}{//\ $\mathbf{R}\in\mathbb{R}^{n\times h\times \frac{d}{h}}$}
    \STATE{\textbf{Return} $\mathbf{R}$}
\end{algorithmic}
\end{algorithm}

\subsection{Flowformer Architecture}

Suppose the model contains $L$ layers with length-$n$ input $\mathbf{X}\in\mathbb{R}^{n\times d_{\text{input}}}$. The overall equations of the $l$-th layer are:
\begin{equation}\label{equ:overall_equation}
  \begin{split}
    \mathbf{Z}^{l}&=\mathrm{Layer}\text{-}\mathrm{Norm}\Big(\mathrm{Flow}\text{-}\mathrm{Attention}(\mathbf{X}^{l-1},\mathbf{X}^{l-1},\mathbf{X}^{l-1})+\mathbf{X}^{l-1}\Big) \\
    \mathbf{X}^{l}&=\mathrm{Layer}\text{-}\mathrm{Norm}\Big(\mathrm{Feed}\text{-}\mathrm{Forward}(\mathbf{Z}^{l})+\mathbf{Z}^{l}\Big), \\
  \end{split}
\end{equation}
where $\mathbf{X}^{l}\in\mathbb{R}^{n\times d}, l\in\{1,\cdots,L\}$ denotes the output of the $l$-th layer with $d$ channels. The initial input $\mathbf{X}^{0}=\mathrm{Embedding}(\mathbf{X})\in\mathbb{R}^{n\times d}$ represents the embedded raw data. $\mathbf{Z}^{l}\in\mathbb{R}^{n\times d}$ is the $l$-th layer's hidden representation.

\subsection{Experiment Configuration} \label{appendix:exp_details}

For the Long-Range Arena \cite{Tay2021LongRA} and the WikiText-103 \cite{Merity2017PointerSM} benchmarks, we just following the official public protocol in Long-Range Arena\footnote{\url{https://github.com/google-research/long-range-arena}} and Fairseq\footnote{\url{https://github.com/pytorch/fairseq}}. Here are the configurations for the other three benchmarks. All details can be found in our code: \href{https://github.com/thuml/Flowformer}{https://github.com/thuml/Flowformer}.

\paragraph{ImageNet-1K} 
\update{The image with resolution $224\times 224$ is split into several patches by convolution at the beginning. Inside the model, we adopt the convolution layer for down sampling between different stages. Thus, different from the model architecture in ViT \cite{dosovitskiy2021an}, we adopt a hierarchical architecture of Transformer without classification token.} The global information is aggregated with the global average pooling layer in the end with the last projector for classification. The details are summarized in Table \ref{tab:model_arch}.

\begin{table}[h]
\vspace{-15pt}
	\caption{Hierarchical architecture for vision recognization task.}
	\label{tab:model_arch}
	\vskip 0.1in
	\centering
	\begin{small}
		\begin{sc}
			\renewcommand{\multirowsetup}{\centering}
			\scalebox{1}{
			\begin{tabular}{l|ccccccccc}
            \toprule
            Parameters & Stage 1 & Stage 2 & Stage 3 & Stage 4  \\ 
            \midrule
            Layers & 3 & 3 & 10 & 3 \\
            Channels & 96 & 192 & 384 & 786 \\
            Heads & 16 & 16 & 16 & 16 \\
            \midrule
            Input sequence length & 3136 & 784 & 196 & 49 \\
             \bottomrule
			\end{tabular}}
		\end{sc}
	\end{small}
	\vspace{-10pt}
\end{table}

\paragraph{UEA} 
Here is the statistical details of the UEA time series classification dataset. As shown in Table \ref{tab:uea}, this benchmark includes various types of subsets, such as the low or high dimension, long or short length, sufficient or insufficient data, many or few class numbers. Thus, experiments on this benchmark can provide a comprehensive comparison. 

\begin{table}[h]
\vspace{-15pt}
	\caption{Statistical Results of the UEA dataset.}
	\label{tab:uea}
	\vskip 0.1in
	\centering
	\begin{small}
		\begin{sc}
			\renewcommand{\multirowsetup}{\centering}
			\scalebox{1}{
			\begin{tabular}{l|ccccccccc}
            \toprule
            Dataset                 & TrainSize & TestSize & NumDimensions & SeriesLength & NumClasses  \\ \midrule
            EthanolConcentration    & 261       & 263      & 3             & 1751         & 4                   \\ 
            FaceDetection   & 5890      & 3524     & 144           & 62           & 2                   \\ 
            Handwriting             & 150       & 850      & 3             & 152          & 26                  \\ 
            Heartbeat               & 204       & 205      & 61            & 405          & 2                   \\ 
            JapaneseVowels          & 270       & 370      & 12            & 29           & 9                   \\ 
            PEMS-SF                 & 267       & 173      & 963           & 144          & 7                   \\ 
            SelfRegulationSCP1      & 268       & 293      & 6             & 896          & 2                   \\ 
            SelfRegulationSCP2      & 200       & 180      & 7             & 1152         & 2                   \\ 
            SpokenArabicDigits      & 6599      & 2199     & 13            & 93           & 10                  \\ 
            UWaveGestureLibrary     & 120       & 320      & 3             & 315          & 8                   \\ \bottomrule
			\end{tabular}}
		\end{sc}
	\end{small}
	\vspace{-10pt}
\end{table}

\begin{figure*}[h]
\begin{center}
	\centerline{\includegraphics[width=0.7\columnwidth]{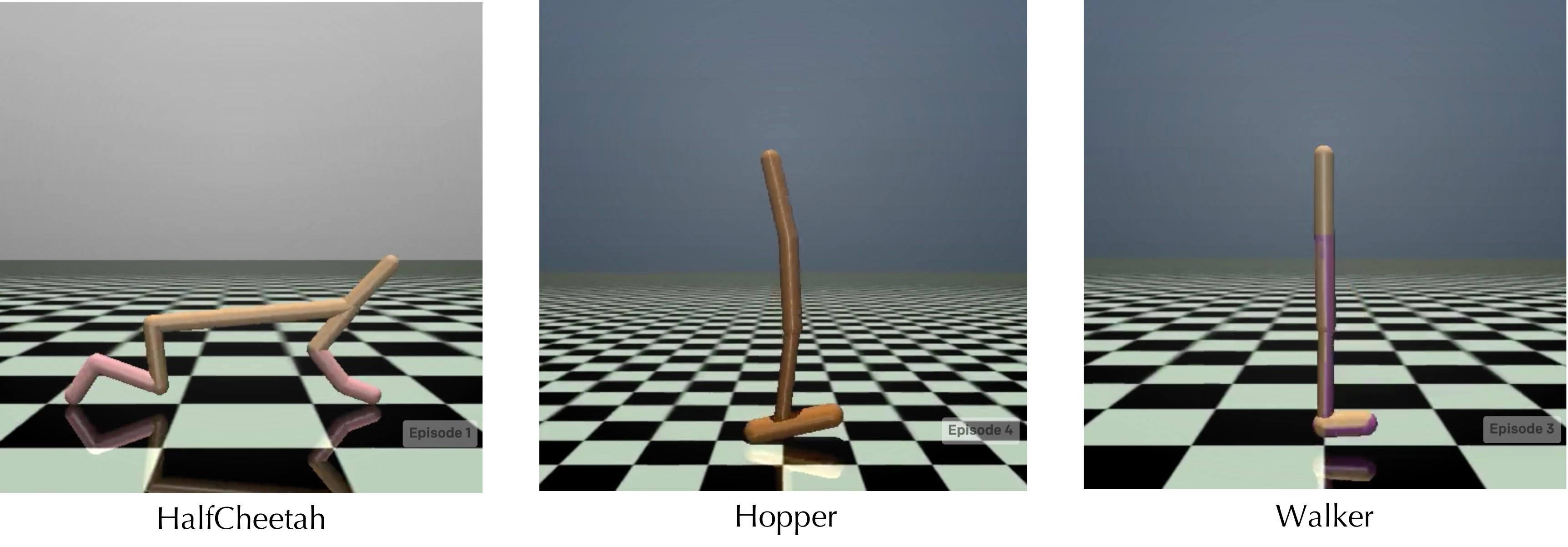}}
	\vspace{-5pt}
	\caption{HalfCheetah, Hopper and Walker environments. }
	\label{fig:rl}
\end{center}
\vspace{-20pt}
\end{figure*}

\paragraph{D4RL} 
This benchmark is for offline RL \cite{lange2012batch, levine2020offline}, which is to learn the policy from a pre-collected dataset and then perform the action in the online environment. This task is challenging not only because of the difficulty of continuous control also due to the extrapolation error caused by the out-of-distribution actions. We experiment on the HalfCheetah, Hopper and Walker environments (Figure \ref{fig:rl}) and train the model with the datasets collected by different policies. We follow the configuration in Decision Transformer \cite{chen2021decisiontransformer} for experiments. 

\section{Ablation studies}

\subsection{Ablation on Activate Function for Non-negative Operation}

As stated in Eq.~\eqref{equ:in-out-flow}, to satisfy the non-negative requirement in flow network, we set the activate function $\phi$ as the sigmoid function. In section, we compare the Flowformer performance under different choices of $\phi$, such as $\text{elu}(\cdot)+1.0$, $\text{Relu}(\cdot)$. As shown in Table \ref{tab:ablation_in_phi}, all the choices of $\phi(\cdot)$ can achieve state-of-the-art performance. Benefiting from the optimization property and numerical stability, the sigmoid function achieves the best averaged performance.

\begin{table*}[h]
	\caption{Ablation results on the Long-Range Arena under different choices of activate function.}
	\vspace{-5pt}
	\label{tab:ablation_in_phi}
	\vskip 0.15in
	\centering
	\begin{small}
		\begin{sc}
			\renewcommand{\multirowsetup}{\centering}
			\setlength{\tabcolsep}{4.3pt}
			\begin{tabular}{l|ccccc|c}
				\toprule
				Model & ListOps $\uparrow$ & Text $\uparrow$ & Retrieval $\uparrow$ & Image $\uparrow$ & Pathfinder $\uparrow$ & Avg $\uparrow$ \\
				\midrule
				$\text{elu}(\cdot)+1.0$ & 38.65 & 64.09 & 61.76 & 43.75 & 71.80 & 56.01 \\
				$\text{Relu}(\cdot)$ & 38.45 & 63.90 & 62.17 & 42.85 &  72.96 & 56.07 \\
				\midrule
				\textbf{Sigmoid function (final version)} & 38.70 & 64.29 & 62.24 & 43.20 & 73.95 & \textbf{56.48} \\
				\bottomrule
			\end{tabular}
		\end{sc}
	\end{small}
\end{table*}

\subsection{Ablation on Activate Functions for Competition and Allocation}

\update{As shown in Eq.~\eqref{equ:overall}, the final design of Flowformer chooses the $\operatorname{Softmax}$ for Competition and the $\operatorname{Sigmoid}$ for Allocation, denoted by $\operatorname{Softmax}$-$\operatorname{Sigmoid}$. Obviously, there are 4 different types of choices. As shown in Table \ref{tab:lra_2x2}, we experiment on every choice exhaustively and find that $\operatorname{Softmax}$-$\operatorname{Sigmoid}$ is the best. These results indicate that in Flow-Attention, $\operatorname{Softmax}$ is more suitable for Competition and $\operatorname{Sigmoid}$ fits the Allocation better. It is because the former is to highlight the tokens and the latter is to control the information flow as stated in the main text.}

\begin{table}[h]
    \vspace{-15pt}
	\caption{Accuracy results (\%) on LRA (averaged from 5 sub-tasks).}
	\vspace{-5pt}
	\label{tab:lra_2x2}
	\vskip 0.1in
	\centering
	\begin{small}
		\begin{sc}
			\renewcommand{\multirowsetup}{\centering}
			\setlength{\tabcolsep}{10pt}
			\scalebox{1}{
			\begin{tabular}{l|cc|c}
				\toprule
			    \scalebox{1}{Model} & \scalebox{1}{Competition} & \scalebox{1}{Allocation} & \scalebox{1}{Avg Acc.} \\
				\midrule
				\scalebox{1}{Choice 1} & \scalebox{1}{$\operatorname{Sigmoid}$} & \scalebox{1}{$\operatorname{Sigmoid}$} & \scalebox{1}{52.36} \\
				\scalebox{1}{Choice 2} & \scalebox{1}{$\operatorname{Sigmoid}$} & \scalebox{1}{$\operatorname{Softmax}$} & \scalebox{1}{53.11} \\
				\scalebox{1}{Choice 3} & \scalebox{1}{$\operatorname{Softmax}$} & \scalebox{1}{$\operatorname{Softmax}$} & \scalebox{1}{55.41} \\
				\scalebox{1}{\textbf{Flowformer}} & \scalebox{1}{$\operatorname{Softmax}$} & \scalebox{1}{$\operatorname{Sigmoid}$} & \scalebox{1}{\textbf{56.48}} \\
				\bottomrule
			\end{tabular}}
		\end{sc}
	\end{small}
	\vspace{-7pt}
\end{table}

\section{Preliminaries of Flow Network}

A flow network \cite{Ahuja1993NetworkF} is a directed graph where each edge has a capacity and each edge receives a flow. The amount of flow on an edge cannot exceed the capacity of the edge. \update{The conservation property can be intuitively explained as follows: for each node, the incoming flow capacity is equal to the outgoing flow capacity.}

\update{In our paper, in addition to considering the inner source-sink flow shown in Figure \ref{fig:flow-view}, we demonstrate that attention mechanisms also exchange information with model's other parts. As presented in Figure \ref{fig:flow_conservation}, \emph{Outside} the attention mechanism, the information of values ($\mathbf{V}$) is obtained from previous layer and the information of results ($\mathbf{R}$) will be provided to the next layer. Without loss of generality, we set the information incoming from or outgoing to other layers as the default value 1. Eq.~\eqref{equ:conservation_proof} of the main text is to prove that, by conducting normalization operations formalized in Eq.~\eqref{equ:conservation}, the incoming flow of each sink and the outgoing flow of each source are equal to 1 respectively, which is exactly the default value that we set for information flow between other layers. Thus, our design follows the flow conservation property well. It is also notable that our design is inspired by flow conservation property and the key insight is ``fixed resource will cause competition".}

\section{More Attention Visualizations}

We provide more attention visualizations on different tasks for a better intuitive understanding.
\subsection{Image Recognition}
\begin{figure*}[h]
\begin{center}
	\centerline{\includegraphics[width=\columnwidth]{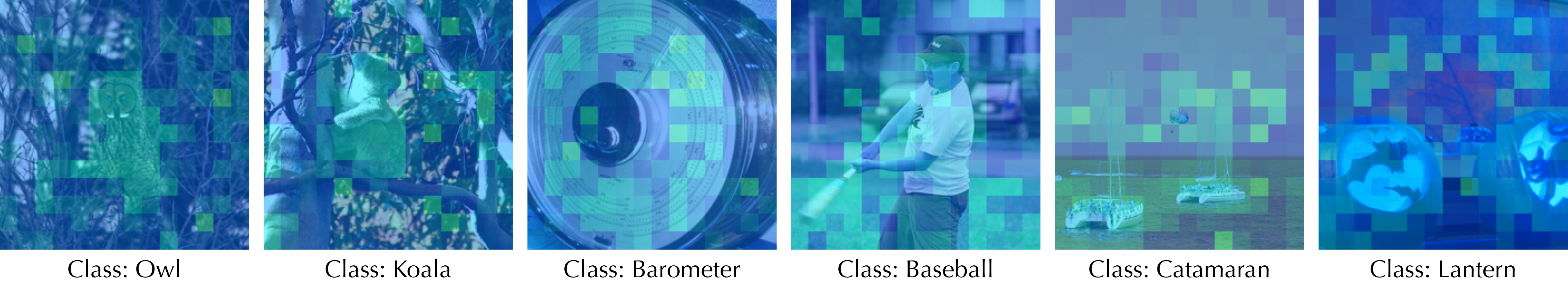}}
	\vspace{-5pt}
	\caption{Visualization of learned allocation weights $\operatorname{Sigmoid}(\widehat{\mathbf{I}})$ of Flowformer. }
	\label{fig:supp_allocation}
\end{center}
\vspace{-25pt}
\end{figure*}

\begin{figure*}[h]
\begin{center}
	\centerline{\includegraphics[width=\columnwidth]{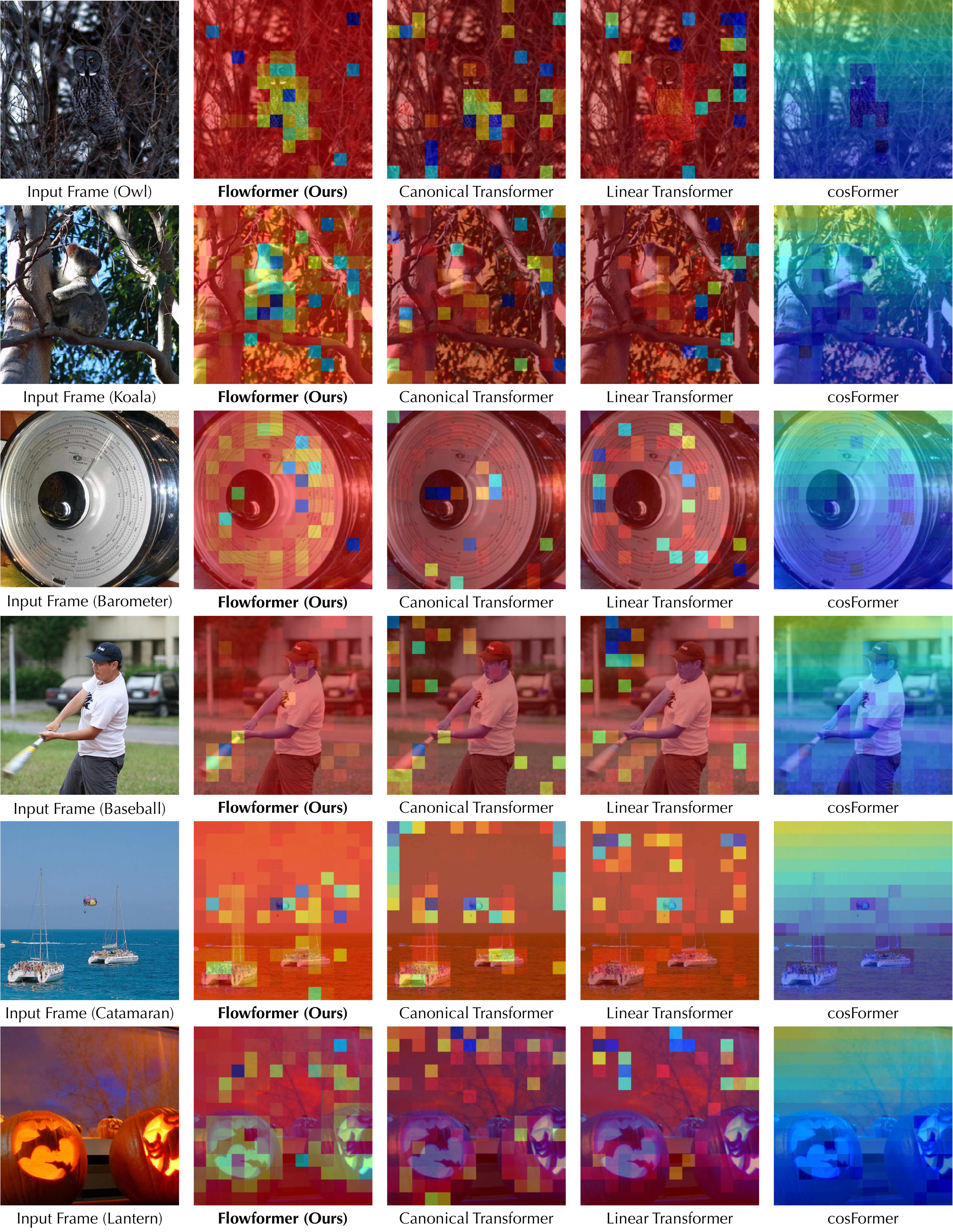}}
	\caption{Comparison of learned attention in the ImageNet-1K dataset. }
	\label{fig:supp_attention}
\end{center}
\vspace{-10pt}
\end{figure*}

In this section, we provide the attention visualization from various classes in the ImageNet-1K. As shown in Figure \ref{fig:supp_attention}, Flowformer can capture the essential part in the image precisely. While the canonical Transformer \cite{NIPS2017_3f5ee243} may appear the distraction problem and Linear Transformer \cite{Katharopoulos2020TransformersAR} could show a trivial attention without the attention concentration. We also visualize the learned allocation weights for corresponding classes in Figure \ref{fig:supp_allocation}, which can also cover the most informative parts for classification.

\subsection{Time Series Classification}\label{appdix:ts_vis}

We provide the attention visualization for the SpokenArabicDigits dataset. This dataset contains times series of mel-frequency cepstrum coefficients (MFCCs) corresponding to spoken Arabic digits from 0 to 9. Note that all the Transformer-based models achieve a great performance in this dataset, including Transformer (\citealt{NIPS2017_3f5ee243}, 98.4\% accuracy), Linear Transformer (\citealt{Katharopoulos2020TransformersAR}, 98.0\%), cosFormer (\citealt{anonymous2022cosformer}, 98.8\%) and Flowformer (ours, 98.8\%). Thus, we provide the visualization only to demonstrate how these attentions work in time series.

\begin{figure*}[h]
\begin{center}
	\centerline{\includegraphics[width=\columnwidth]{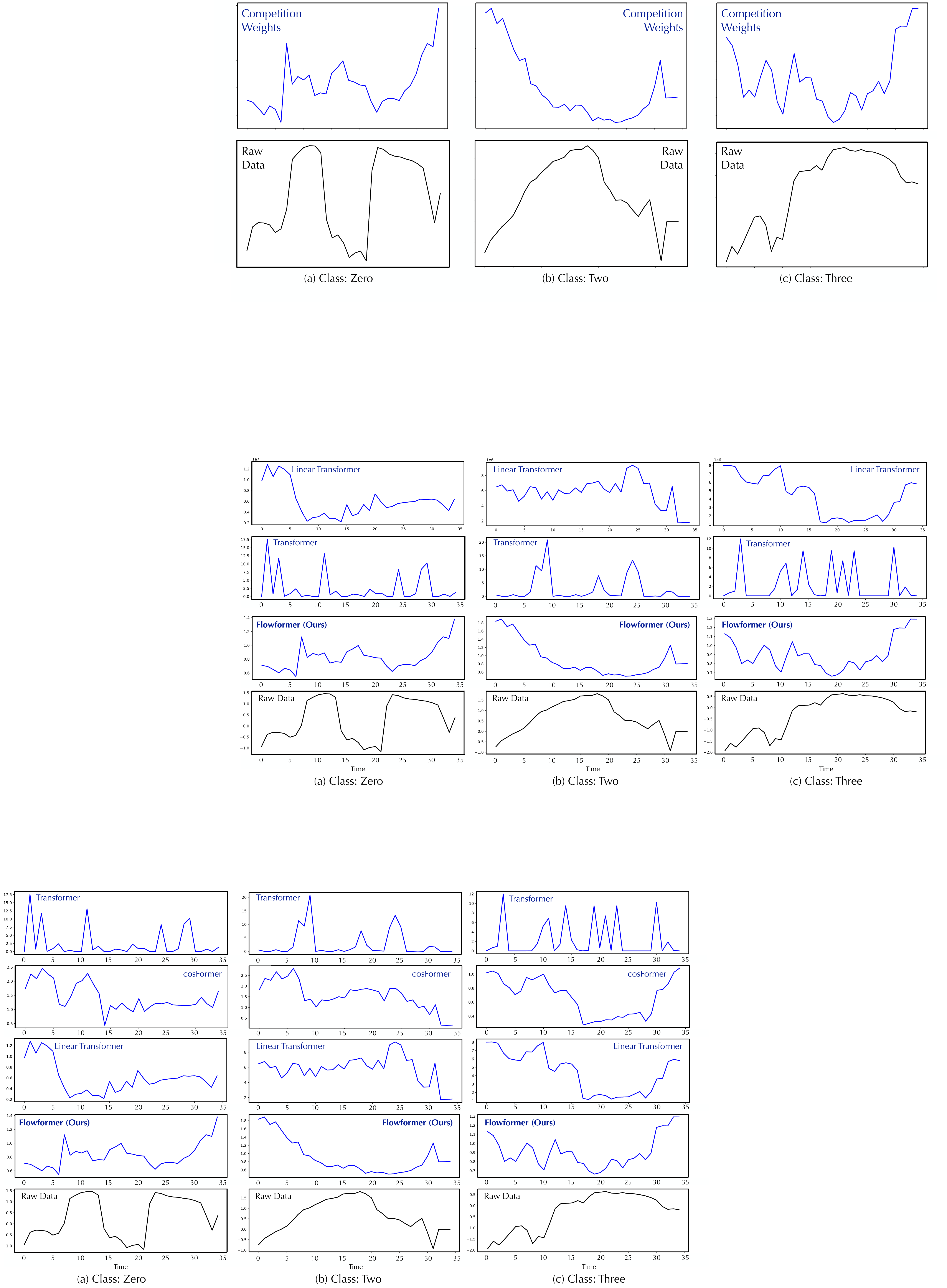}}
	\caption{Learned attention in the SpokenArabicDigits dataset. We plot the competition weights $\operatorname{Softmax}(\widehat{\mathbf{O}})$ for Flowformer.}
	\label{fig:ts_attn}
\end{center}
\vspace{-10pt}
\end{figure*}

Generally, we can find that Flowformer successively pays non-trivial attention to different phases of the time series for different classes (Figure \ref{fig:ts_attn}). Concretely, as shown in Figure \ref{fig:ts_attn}(a), Flowformer pays more attention on the intervals $[5,10]$ and $[30,35]$, where are important for the classification of the two-stage pronunciation for ``zero'' ([\textprimstress ziro\textipa{U}]). Besides, Flowformer pays more attention to the distinguishable part in ``two'' ([tu\textipa{:}], Figure \ref{fig:ts_attn}(b)) and ``three'' ([\textipa{T}ri\textipa{:}], Figure \ref{fig:ts_attn}(c)). While for the vanilla Transformer, it may be distracted by unimportant fluctuations (Figure \ref{fig:ts_attn}(c)). As for the Linear Transformer and cosFormer, they may fail to capture the second phase in the pronunciation of ``zero'' (Figure \ref{fig:ts_attn}(a)).

\end{document}